\documentclass[runningheads]{llncs}
\usepackage{graphicx}
\usepackage{multirow}
\usepackage{amsfonts}
\usepackage{epsfig}
\usepackage{amsmath}
\usepackage{amssymb}
\usepackage[nolist]{acronym}
\usepackage[english]{babel}
\usepackage{cite}
\usepackage{color}
\usepackage{stfloats}
\usepackage{algorithm}
\usepackage{algorithmic}
\usepackage{subfigure}
 \usepackage{soul}
\usepackage{psfrag}

\newcommand{\lt}[1] {\{\textbf{\textcolor[rgb]{1.0,0.0,0.0}{LT: }}\textit{{#1}}\}}
\newcommand{\pfnote}[1] {\{\textbf{\textcolor[rgb]{0.0,0.0,1.0}{PF: }}\textit{{#1}}\}}

\newcommand{\algo}{\textcolor[rgb]{0.76,0.5,0.18}{\selectfont\ttfamily{{Grab-UCB}}}}
\newcommand{\algolight}{\textcolor[rgb]{0.76,0.5,0.18}{\selectfont\ttfamily{{Grab-arm-Light}}}}

\newcommand{\mat}{\pmb}

\begin{document}
\begin{acronym}
\acro{SIC}{successive interference cancellation}
\acro{BN}{burst node}
\acro{SN}{slot node} 
\acro{MAB}{Multi-arm bandit} 
\acro{DMS}{decision making strategies}
\acro{GSP}{Graph Signal Processing}
\acro{RKHS}{Reproducing kernel Hilbert space}
\end{acronym}

\title{Online Network Source Optimization with Graph-Kernel MAB}

\titlerunning{Grab-UCB}
%
\author{Laura Toni\footnote{Corresponding author: l.toni@ucl.ac.uk}\inst{1} \and
Pascal Frossard\inst{2}}
\authorrunning{Toni, Frossard}

\tocauthor{Laura Toni, Pascal Frossard}
\toctitle{Online Network Source Optimization with Graph-Kernel MAB}

\institute{  
EEE Dept, University College London, London, United Kingdom \and LTS4, EPFL, Lausanne, Switzerland \\
}

%

%
\maketitle              
\begin{abstract}
We propose \algo, a graph-kernel multi-arms bandit  algorithm to  learn  online the optimal source placement in large scale networks, such that the reward obtained from a priori unknown network processes is maximized.  The uncertainty calls for online learning, which suffers however from the curse of dimensionality. To achieve sample efficiency,   we describe the network processes with an adaptive graph dictionary model, which typically leads to sparse spectral representations. This enables a data-efficient learning framework, whose learning rate scales with the dimension of the spectral representation model instead of the one of the network. We then propose \algo, an online sequential decision strategy that learns the parameters of the spectral representation while optimizing the action strategy. We derive the performance guarantees that depend on network parameters, which further influence the learning curve of the sequential decision strategy 
We  introduce a computationally simplified solving method, \algolight, an algorithm  that walks along the edges of the polytope representing the objective function. Simulations results show that the proposed online learning algorithm outperforms baseline offline methods that typically separate the learning phase from the testing one. The results confirm the theoretical findings, and further highlight the gain of the proposed online learning strategy in terms of cumulative regret, sample efficiency and computational complexity.

\end{abstract}


\section{Introduction}
\label{sec:intro}

Large-scale interconnected systems (transportation networks, social networks, etc.), which create services and produce massive amounts of data, are becoming predominant in many application domains. The management of such networked systems is exceedingly hard because of their intrinsic and constantly growing complexity. Many works have been proposed to tackle this problem (e.g., model based optimal control, consensus works~\cite{Zhang:J15, Movric:J14, Salami:J17,Yuan:A17,Nassif:A19}, Bayesian approaches \cite{Acemoglu2011}, etc.) but with a limited focus on online learning and control of large-scale networks. The latter becomes extremely challenging with dynamic and high dimensional network processes that controls the evolution of states of network nodes. The dynamics introduce uncertainty about the system environment, which can be addressed by online learning strategies that infer the system behaviour before taking the appropriate adaptation actions or decisions. The high dimensionality of the problem calls for proper information representation methods.

We consider the particular problem of optimal source placement in order to maximize a reward function on a network, which depends on network processes that are a priori unknown and must be learned online. We address this challenge by blending together online learning theory~\cite{lattimore2018} and \ac{GSP} \cite{shuman2013emerging,Ortega:J18} with the key intuition that the latter permits to appropriately model the large-scale network processes via sparse graph spectral representations. This generates a data-efficient learning framework, whose learning rate does not scale with the dimension of the network as in most methods of the literature, but rather with the dimension of the spectral representation. Indeed, in classical online learning solutions such as those casted as \ac{MAB} problems, the main learning steps  happen in the action (or node) domain and do not scale properly with the search space. The key intuition underpinning our new framework is to consider these learning steps at the crossroad of the search space (or node) domain  and the latent space (or spectral) domain. An agent takes sequential decision strategies in the high-dimensional vertex domain based on the uncertainty of the model estimated in the low-dimensional spectral domain. Similar intuition is shared in literature on bandit for spectral graph domains (see Sec.~\ref{sec:related works}), which aims at applying linUCB algorithms while preserving the low-dimensionality assumption of the reward. However, we do not limit ourselves to smoothness assumption in inductive bias and rather model the entire network process as a graph filter that is excited by a set of unknown low-rank inputs (action). 
As a result, the learning process boils down to inferring spectral graph representations with a learning rate that scales with the dimension of their generating kernels, which is substantially lower than the one of the search space. 
With our framework, this online learning problem can be reformulated and reduced to a linearUCB problem \cite{chu2011contextual} that is a well-known algorithm achieving good sample efficiency in the literature for linear stochastic bandit problems.  We then derive the theoretical bound of the estimation of the graph spectral model and translate it to the MAB upper confidence bound. Finally, we observe that the optimization  method leads to an arm selection problem that is NP-hard, and  we provide a low-complexity algorithm, \algolight, by exploiting the structure of the optimization function (maximization of a convex function over a polytope). Simulations validate the accuracy of the proposed low-complexity algorithm as well as the gains of the proposed graph-kernel MAB strategy, in terms of cumulative regret, sample efficiency and computational complexity, when compared to baseline offline methods.

The reminder of this paper is as follows. The online source optimization problem is formulated in Sec.~\ref{sec:source_optimization_problem}. The proposed  \algo \, is detailed in Sec.~\ref{sec:Online_Source_Optimisation},  and the low complexity solution \algolight \, is introduced in Sec.~\ref{sec:solving_method}. Simulation results  
are discussed in Sec.~\ref{sec:results}. Related works are presented in Sec~\ref{sec:related works}, and conclusions in Sec.~\ref{sec:conclusion}.

\section{Online Source Optimization Problem}
\label{sec:source_optimization_problem}
\subsection{Problem Formulation}
 \label{sec:prob_form_new}
\begin{figure*}[!t]
\centering
\includegraphics[width=.8\linewidth,  draft=false]{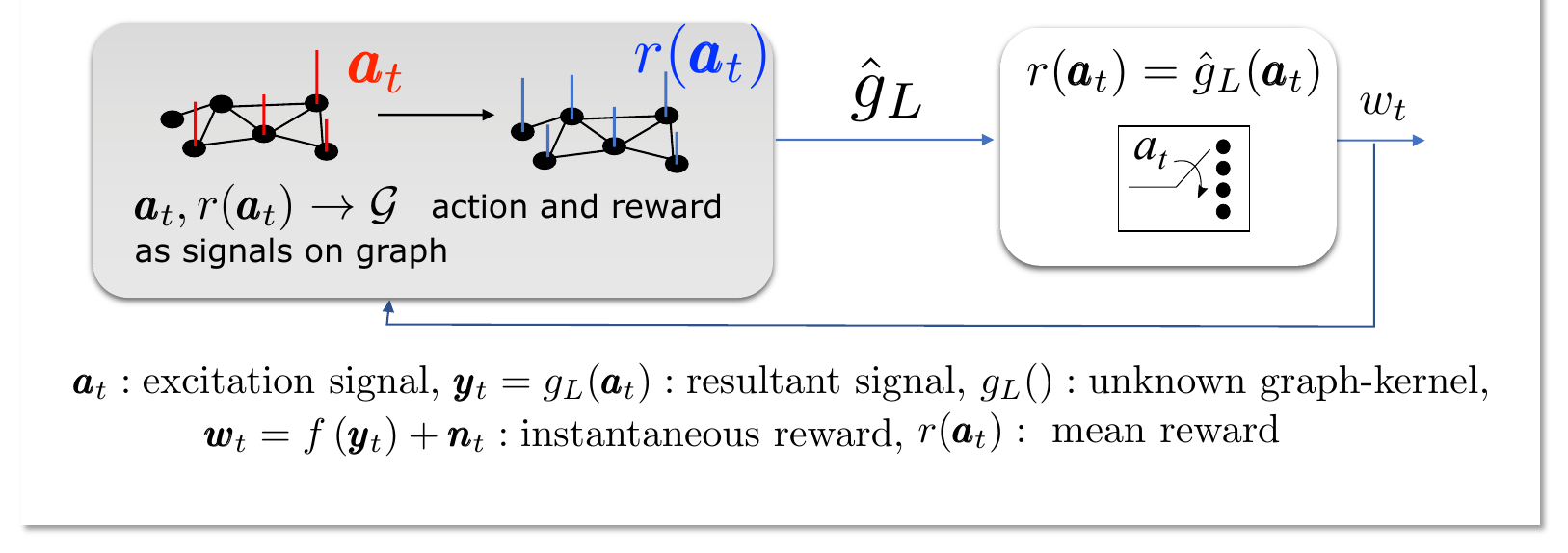} 
\caption{Graphical visualisation of the proposed framework. } \label{fig:proposed_framework}
\end{figure*}

Let consider a learner (or agent) controlling processes on large scale networks with no \textit{a priori} information on their dynamics. Examples can be  network cooling systems~\cite{jones2018:nature}, opinions spreading across social networks~\cite{perra2019modelling}, or energy distribution networks~\cite{ramakrishna2021grid} that need to be managed online with no a priori information about the underlying processes.  In this paper, we model these processes as signals on graphs, as depicted in Fig. \ref{fig:proposed_framework}, with actions and resultant signal defined on the weighted and undirected graph $\mathcal{G}=(\mathcal{V},\mathcal{E}, \mat{W})$ with $\mathcal{V}$ being  the vertex set ($|\mathcal{V}|=N$), $\mathcal{E}$ the edge sets, and \mat{W} the $N \times N$ graph adjacency matrix.   
Namely, we assume that the  action taken by the learner at time $t$, $\mat{a}_t\in\mathbb{R}^N$, is modelled as an \emph{excitation signal} on the graph and produces a \emph{resultant graph signal} $\mat{y}_t\in\mathbb{R}^N$. The  instantaneous reward $\mat{w}_t = \mathcal{V} \rightarrow \mathbb{R}^N$ can be modelled as a function of the resultant graph signal, and reads $\mat{w}_t = f\left(\mat{y}_t\right) + \mat{n}_t $, with $f(\cdot)$ being an affine function\footnote{This includes many reward shapes such as subsampled or filtered signal as well as mean value.} and $\mat{n}_t $ an additive noise. The overall goal of the agent is to learn which actions need to be selected to achieve the maximum reward, with no prior information on the network process (i.e., the mapping from $\mat{a}_t$ to $\mat{w}_t$).  The problem can be casted as a stochastic MAB problem, aimed  minimizing the cumulative loss (or equivalently maximize the cumulative reward) over a time horizon $T$, which is seen as the minimization of the pseudo regret $R_T= T r(\mat{a}^{\star}) - \sum_{t=1}^T r(\mat{a}_t)$, with  $r(\mat{a})$ being the mean reward for action $\mat{a}$.

In this following,  we consider the online source optimization problem\footnote{It is worth noting that the formalism introduced in this Section extends to most problem on learning on network process, but for the sake of brevity and clarity we discuss only the source optimization problem.}, where a decision maker needs to select $T_0$ sparse actions out of $N$, i.e., $ \mathcal{A}  = \{  \mat{h} \  \vert   \ ||\mat{h} ||_0 \leq T_0  \  \wedge   \  h_{n} \in [0,1],  \ n=1, . . . , N   \}$, where $T_0$ is the maximum sparsity level of the actions. We assume that the rewards associated to consecutive actions are independent and model the affine function $f(\cdot)$ as the mapping function $\mat{M}\in\mathbb{R}^{N\times N}$, a diagonal binary matrix, with the $n$-th diagonal element being $1$ if the signal at the node $n$ is consdiered in computing the reward, or $0$ if that signal is masked. Real-world examples are influence maximization problems, such as targeted advertisement online~\cite{tang2018social,ide2022targeted} or optimization of cooling systems and/or power networks~\cite{datacenters}.
This online learning problem can be solved by classical MAB problems, with a sublinear regret  $R_T =\mathcal{O}(|\mathcal{A}|\log T)$~\cite{lattimore2018}, with $|\mathcal{A}|={N \choose T_o}$, if $T_0$ is the imposed sparsity of $\mat{h}$. This regret is not sustainable in large-scale networks  with large  action space $|\mathcal{A}|$. 


\subsection{Graph-Kernel MAB Framework}
\label{sec:graph_kernel_MAB}
We now propose a graph-kernel MAB problem that exploits the geometry of the network processes to achieve a better regret scaling. Specifically, we  model the mapping $\mat{a}_t \rightarrow \mat{w}_t$ as an \emph{unknown} structured function of the graph Laplacian $\mat{L}$ (defined to be $\mat{L}=\mat{D}-\mat{W}$, with $\mat{D}$ being the degree matrix), i.e.,  \begin{align}\label{eq:reward}
\mat{w}_t = f\left(\mat{y}_t\right)+ \mat{n}_t =  f\left({g}_L(\mat{a}_t)\right)+ \mat{n}_t    
\end{align}
with ${g}_L(\cdot)$ being an \emph{unknown} generating kernel\footnote{Graph filter defined in the spectral domain of the graph, typically in the form of the power series of the graph Laplacian\cite{Thanou:J14}.} of the graph Laplacian $\mat{L}$. The generating kernel models the process on graphs and characterizes the effect of an action in a resulting graph signal, which will impact the mean reward. Hence, the agent infers the mapping $\mat{a}_t \rightarrow r\left({\mat{a}_t}\right)$ by learning the graph generating kernel ${g}_L(\cdot)$ in the spectral domain, which is much more sample-efficient than learning the mapping $\mat{a}_t \rightarrow r\left({\mat{a}_t}\right)$ directly in the high-dimensional vertex (action) space.  
 
We formulate the online learning problem via GSP tools, and we cast the problem into a linear MAB problem, in which the confidence bound is defined on the graph spectral parameters of the generating kernel. We model the network process via the graph-based parametric dictionary learning algorithm in~\cite{Thanou:J14}, with a signal on graph defined as   $\mat{y} =\mat{\mathcal{D}}\mat{h}+ \mat{\epsilon}$,  with $ \mat{h}= [h_{1}, h_{2}, \ldots,  h_{N}]^T$ being the latent variables (localized events) defined on the graph, i.e., the excitation signal defined as actions in our model, and $\mat{\epsilon}=[\epsilon_{1}, \epsilon_{2}, \ldots, \epsilon_{N}]^T$ is a Gaussian and $N$-dimensional random variable with $ \epsilon_{n}\sim\mathcal{N}(0,\sigma_{e}^2)$~\cite{chu2011contextual}.

The graph dictionary $\mat{\mathcal{D}}$ is defined as $\mat{\mathcal{D}} = g_{\mat{L}}(\cdot)= \sum_{k=0}^{K-1} \alpha_{k} \mat{L}^k$~\cite{Thanou:J14} and represents the graph-kernel, which incorporates the intrinsic geometric structure of data domain into the atoms of the dictionary through $\mat{L}$. Assuming that signals have a support contained within $K$ hops from vertex $n$,    
the resulting signal in $n$ can  be represented  as combinations of localized events (e.g., local signals) on the graph, which can appear in different vertices and diffuse along the graph. Namely, 
\begin{align}
y_{n} = \sum_{m=1}^N h_{m} \sum_{k=0}^{K-1} \alpha_k (\mat{L}^k)_{n,m} + \epsilon_{n} 
\end{align}
where $(\mat{L}^k)_{n,m}$ is the $(m,n)$ entry of $\mat{L}^k$ and we recall that $(\mat{L}^k)_{n,m}=0$ if the shortest path between $n$ and $m$ has a number of hops that is greater than $k$. With the following matrix notations where $\mat{P}=[\mat{L}^0, \mat{L}^1, \mat{L}^2, \ldots, \mat{L}^{K-1}]$, with $\mat{P}\in\mathbb{R}^{N\times NK}$, captures the powers of the Laplacian, and with $\mat{\alpha=}[\alpha_0, \alpha_1, \ldots, \alpha_{K-1}]^T$ representing the polynomials coefficients in the dictionary, we can rewrite the resulting signal as
\begin{align}\label{ee:signal}
\mat{y} = g_L(\mat{h}; \mat{\alpha}) = \mat{P} \mat{I}_{K} \otimes \mat{h}  \mat{\alpha} + \mat{\epsilon}   = \mat{PH}  \mat{\alpha}  + \mat{\epsilon}
\end{align}
with $\mat{I}_{K}$ being the $K \times K$ identity matrix,  $\otimes$ the Kronecker product, and $\mat{H} =\mat{I}_{K} \otimes \mat{h}$, with $\mat{H}\in\mathbb{R}^{NK\times K}$. In our framework, at the decision opportunity $t$,  the agent controls the latent variables $\mat{h}_t$ while learning the polynomial coefficients $\mat{\alpha}$.  Given that the instantaneous reward is an affine function of  the resultant signal $\mat{y}_t$,
substituting \eqref{ee:signal} in the reward expression \eqref{eq:reward}, we achieve the following 
\begin{align}\label{eq:reward}
 \mat{w}_t &=  \mat{M}  \mat{y}_t   = \mat{M}  \mat{PH_t}  \mat{\alpha}+ \mat{M} \mat{\epsilon}_t = \ \mat{X} \mat{\alpha}   + \mat{n}_t
\end{align}
where  and $\mat{X} = \mat{M}  \mat{PH_t}$, with   $\mat{X}_t\in\mathbb{R}^{N\times K}$. 

In short, the reward can be expressed as a linear combination of the $K$-degree polynomial $\mat{\alpha}$  and the matrix $\mat{X}_t$, which includes both the graph structure information (via the Laplacian $\mat{L}$) and the action $\mat{h}_t$.    This is important because: 
\begin{itemize}
   \item the reward is a linear mapping between the unknown parameters $\mat{\alpha}$ and the actions $\mat{\mat{h}_t}$ (hence $\mat{X}_t$), implying that we can solve the online learning problem with the linUCB~\cite{chu2011contextual} theory.  
    \item  the reward is given by the generating kernel $g_L(\cdot)$, which is parametrized by the vector $\mat{\alpha}$ with dimensionality $K$.  Therefore the uncertainty bound in the linUCB is evaluated in the spectral (low-dimensional) domain.  This presents an important advantage, as the regret scales as $\mathcal{O}(d \sqrt{T} \log T)$ in LinUCB, where $d$ is the dimension of the unknown (low-dimension) parameter $\mat{\alpha}$. 
\end{itemize}

\section{\algo: Proposed Algorithm}
\label{sec:Online_Source_Optimisation}

\begin{figure*}[t]
\centering
\includegraphics[width=.8\linewidth,  draft=false]{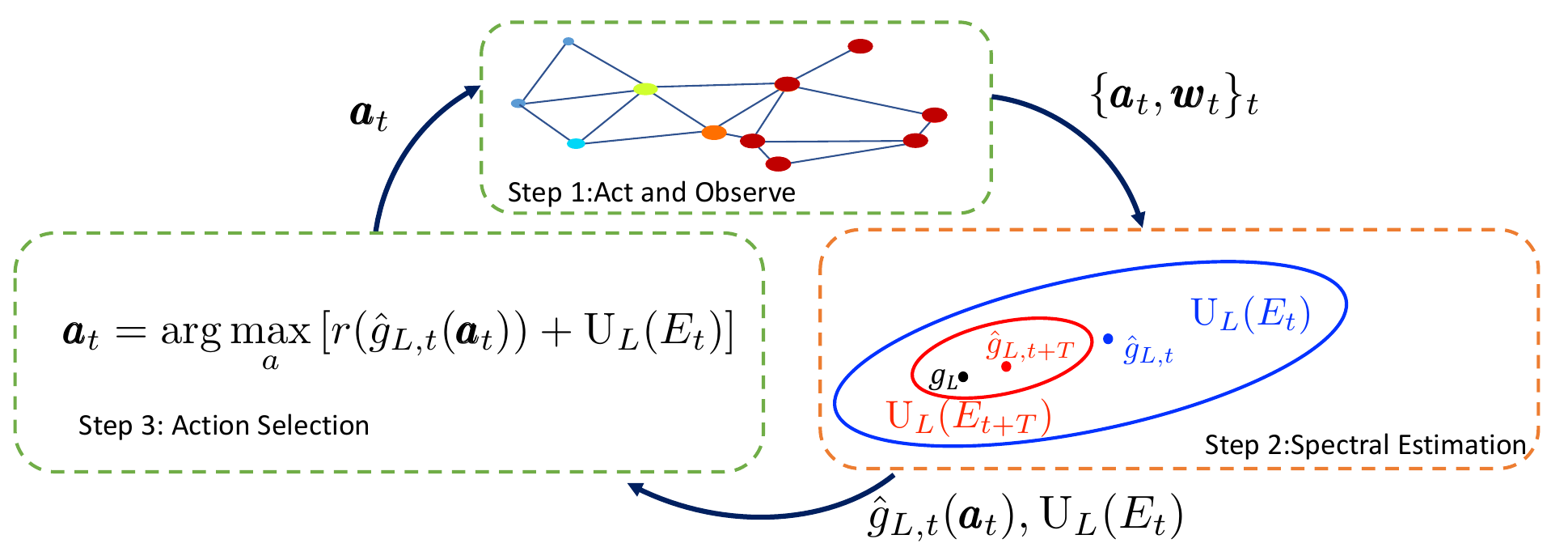} 
\caption{Figurative example of the online graph-strcutured processing. Green (red) dashed boxes are defined in the vertex (spectral) domain.  } \label{fig:Spectral_Learning_fig}
\end{figure*}

We now propose a theoretical bound and algorithmic solution to the online source optimisation problem using the new framework described in  Sec.~\ref{sec:source_optimization_problem}, which permits to learn in the spectral domain and to act in the vertex domain, see Fig.~\ref{fig:Spectral_Learning_fig}.
There are two interacting subtasks in the algoirthm: 1) refinement of the coefficients estimate, 2) selection of the arm given the updated knowledge of the system.

\paragraph{Step 1: Coefficients estimation}

Let consider the $t$-th decision opportunity, when $t-1$ decisions have already been taken and the corresponding signals and rewards have been observed. The training set is built over time thus it corresponds to sequence of pairs  $\{(\mat{h}_{\tau},{\mat{w}}_{\tau})\}_{\tau=1}^{t-1}$, where we recall that $p(\mat{y}| \mat{h},\mat{\alpha})\sim \mathcal{N}(g_L(\mat{h}; \mat{\alpha}),\sigma_e^2 \mat{I}_N)$, and that the randomness is due to the random noise $\mat{\epsilon}_{\tau}$. For large $t$, maximizing the MAP probability $p(\mat{\alpha}|\mat{y},\mat{h})$ corresponds to minimizing the  $l^2$-regularized least-square estimate of $\mat{\alpha}$, leading to the following  problem:
$\hat{\mat{\alpha}}_t: \arg\min_{\mat{\alpha}}  \sum_{\tau=1}^{t-1} || \mat{M} \mat{PH_{\tau}\alpha} -   {\mat{w}}_{\tau}||^2_2 + \mu ||\mat{\alpha}||^2_2\,.$
It follows that 
\begin{align}\label{eq:LSE_solution}
\hat{\mat{\alpha}}_t 
=   \left[ \sum_{{\tau}=1}^{t-1} \mat{Z}_{\tau}^T \mat{Z}_{\tau} + \mu \mat{I}_{K}  \right]^{-1}   \sum_{{\tau}=1}^{t-1} \mat{Z}_{\tau}^T {\mat{w}}_{\tau} 
=    \left[  \mat{Z}_{1:t}^T \mat{Z}_{1:t} +  \mu \mat{I}_{K} \right]^{-1}  \mat{Z}_{1:t}^T  {\mat{W}}_t
=  \mat{V}_t^{-1} \mat{Z}_{1:t}^T {\mat{W}}_t
\end{align}
with $\mat{Z}_{1:t}= [\mat{Z}_1, \mat{Z}_2, \ldots, \mat{Z}_{t-1}]^T$,  $ \mat{Z}_{\tau} =  \mat{M} \mat{PH}_{\tau}$,  ${\mat{W}}_t= [{\mat{w}}_1, {\mat{w}}_2, \ldots, {\mat{w}}_{t-1}]^T$,   and $\mat{V}_t= \mat{Z}_{1:t}^T \mat{Z}_{1:t} +  \mu \mat{I}_{K}$.   
The $l^2$-regularized least-square estimate  leads to an approximation of the actual polynomial $\mat{\alpha}$, and this approximated estimate $\hat{\mat{\alpha}}_t$  is refined at each decision opportunity.

\paragraph{Step 2: Action selection}

Once the estimation of the $\mat{\alpha}$ coefficients is refined, the decision maker needs to select the best action to take for the $t$-th decision opportunity. Following the theory of linear UCB \cite{chu2011contextual}, the decision maker evaluates   the  confidence bound $E_t$ as an ellipsoid centered in $\hat{\mat{\alpha}}_t$ defined such that  $\mat{\alpha} \in E_t$ with probability $1-\delta$ for all $t\geq 1$, see Fig.~\ref{fig:Spectral_Learning_fig}. Then, the decision maker selects the best action that maximizes the estimated  mean reward, for each possible generating kernel in the ellipsoid (optimism in face of uncertainty~\cite{lattimore2018}). Formally, the decision maker selects the action $\mat{h}$ (and therefore $\mat{X} = \mat{M}\mat{P} \mat{I}_{K} \otimes \mat{h}$) such that 
\begin{align}\label{eq:step2}
\mat{h}_t: \arg\max_{\mat{h}\in{\mathcal{A}}} \max_{\mat{\alpha}\in E_t}   \mat{X \alpha}\,.
\end{align}
To apply \eqref{eq:step2}, we need to formally  derive the confidence bound  $E_t$. This can be derived by the following two Lemmas (proofs in Appendix \ref{sec:appendixA}).   Lemma 1 bounds the matrix $\mat{V}_t$, which defines the regularized least-square solution as shown in~\eqref{eq:LSE_solution}.  Lemma 1 is key to evaluate the upper confidence bound in Lemma 2. Specifically, Lemma 2 provides the confidence bound $E_t$ such that $E_t : \{ ||\hat{\mat{\alpha}}_t - \mat{\alpha}_* || \leq c_t \}$.   It is worth noting that both bounds have explicit dependency on topological features of the graph, such as the sum of eigenvalues power, as we comment later. 

 \begin{algorithm}[!t]
  \caption{\algo}
  \label{alg:exact}
  \begin{algorithmic}
\STATE     {\bf Input:}  
\STATE      $N$: nr of nodes, $T_0$: sparsity level of action signal $\mat{h}$, $K$: sparsity of the basis coefficients
\STATE     $\mu, \delta$: regularization and confidence parameters
\STATE     $R, S$: upper bounds on the noise and $\mat{\alpha}_*$
\STATE    $t=1$
\WHILE {$t\leq T$} 
        \STATE  Refine estimate of the coefficients      
        \STATE $\mat{X}_{1:t}= [\mat{X}_1, \mat{X}_2, \ldots, \mat{X}_{t-1}]^T$
        \STATE $\mat{Y}_{1:t}= [\mat{y}_1, \mat{y}_2, \ldots, \mat{y}_{t-1}]^T$
        \STATE  $\mat{V}_t= \mat{X}_{1:t}^T \mat{X}_{1:t} +  \mu \mat{I}_{K+1}$  
        \STATE Step 1: Coefficients estimation: 
        \STATE \ \ \ \  \ $\hat{\mat{\alpha}}_t =   \mat{V}_t^{-1} \mat{X}_{1:t}^T \mat{Y}_{1:t}$ 
        \STATE Step 2: Action Selection
        \STATE \ \ \ \  \    Evaluate the confidence bound and select the best action
	\STATE \ \ \ \  \  Select action by solving \eqref{eq:opt_step2} numerically or via \algolight 
	\STATE  Observe the resulting signal $\mat{y}_t$ and the instantaneous reward  
        \STATE  $t=t+1$
\ENDWHILE 
\end{algorithmic}
\end{algorithm}

\vspace{.3cm}

\emph{{\bf Lemma 1:} Suppose $\mat{Z}_1,  \mat{Z}_2, \ldots,  \mat{Z}_t \in \mathbb{R}^{1\times K}$, with $\mat{Z}_{\tau}=\mat{M}\mat{P}\mat{I}_{K} \otimes \mat{h}_{\tau}$ and for any $1\leq \tau \leq t-1$, $||h_{\tau}||_F^2\leq T_0$, and $||\mat{M}||_F^2\leq Q$. Let $\mat{V}_t = \sum_{\tau} \mat{Z}_{\tau}^T \mat{Z}_{\tau} + \mu \mat{I}_{K}$ with $\mu>0$,  then}
$
|\mat{V}_t| \leq \left[ \mu +   d QT_0 \right]^{K},
$
\emph{with $d=\sum_k \sum_l \lambda_l^k$, with $\lambda_l$ being the $l$-th eigenvalue of the graph Laplacian. }

\vspace{.3cm}

\emph{{\bf Lemma 2:} Assume that $\mat{V}_t = \sum_{\tau} \mat{Z}_{\tau}^T \mat{Z}_{\tau} + \mu \mat{I}_{K}$, define ${\mat{w}}_{\tau} = \mat{Z}_{\tau} \mat{\alpha_*} + \mat{\eta}_{\tau}$, with $\mat{Z}_{\tau}=\mat{M}\mat{P}\mat{I}_{K} \otimes \mat{h}_{\tau}$  and with $\mat{\eta}_t$ being conditionally $R$-sub-Gaussian, and assume that $||\mat{\alpha}_*||_2\leq S$, and $||\mat{h}_{\tau}||_F^2\leq T_0$. Then, for any $\delta>0$,  with probability at least $1-\delta$, for all $t\leq0$, $\mat{\alpha}_*$ lies in the set }
 $$
E_t: \left\{ {\mat{\alpha}} \in \mathbb{R}^{1\times K} :   || \hat{\mat{\alpha}}_t - \mat{\alpha}||_{\mat{V}_t}  \leq   R\left[ \sqrt{K \log(\mu+tdQT_0 ) +  2 \log (\mu^{-1/2}\delta) } \right] + \mu^{1/2}S  
\right\}
$$
\emph{with $d=\sum_k \sum_l \lambda_l^k$, with $\lambda_l$ being the $l$-th eigenvalue of the graph Laplacian and $\hat{\mat{\alpha}}_t$ is the $l^2$-regularized least-square estimate of $\mat{\alpha}$ with $t$ training samples. }
\  
\vspace{.3cm}   \\
From Lemma 2, the   maximization in \eqref{eq:step2} becomes (see Appendix~\ref{sec:bound_variance} for details)
\begin{align}\label{eq:opt_step2}
\mat{h}_t &= \arg\max_{\mat{h}\in{\mathcal{A}}} \max_{\alpha\in E_t}   \mat{X \alpha}   =  \arg\max_{\mat{h}\in{\mathcal{A}}}  \mat{X \alpha} +    c_t\sqrt{\mat{X}\mat{V}_t^{-1}\mat{X}^T}   \nonumber \\
&= \arg\max_{\mat{h}\in{\mathcal{A}}}    \mat{X \alpha} + c_t ||\mat{X}||_{\mat{V}_t^{-1}} = \arg\max_{\mat{h}\in{\mathcal{A}}}   \left[\mat{M}\mat{PH}\hat{\mat{\alpha}}_t  + c_t || \mat{M}\mat{PH}||_{\mat{V}_t^{-1}}\right]\; 
\end{align}
with  $c_t = R\left[ \sqrt{K \log(\mu+tdQT_0 ) +  2 \log (\mu^{-1/2}\delta) } \right] + \mu^{1/2}S $ following Lemma 2. 
This optimization   characterizes the Step 2, \emph{i.e.}, the action selection. However, this cannot be solved efficiently in large scale networks, see Appendix~\ref{appendix:solving_method} for description of the solving method and scalability issues.  In the following Section, we propose a computationally effective optimization algorithm.    \newline

In Algorithm 1, we summarize the main steps of the proposed \algo \ strategy. This algorithm achieves the following regret bound (derived in Appendix~\ref{sec:bound_variance}) of  
 $R_T  \leq 2(c_T +1) \sqrt{  2KT \log \left( 1+ \frac{Q T_0 d}{\mu}  \right)  }$. 
It is worth noting the dependency on the topological structure via   $d$, sum of eigenvalues power.   Finally, the regret does not depend on the network size $N$ but rather on the sparsity level $T_0$, hence the strong gain with respect to linUCB like algorithms. 


\section{\algolight: Efficient Action Selection}
\label{sec:solving_method} 
The methodology proposed in the previous Section entails two main optimization/learning steps that need to be solved. While the solution to the optimization in Step 1 has a closed form, i.e., equation \eqref{eq:LSE_solution}, in Step 2 the optimization problem in \eqref{eq:opt_step2} needs to be solved efficiently. This optimization becomes computationally expensive in large-scale graphs, see Appendix~\ref{appendix:solving_method}. Therefore, in the following we propose a computationally effective optimization algorithm to address the scalability issue that we call \algolight. It is a computationally light solving method aimed at selecting the action/arm in Step 2 of \algo. 
We first rewrite the problem \eqref{eq:opt_step2} (see Appendix~\ref{appendix:solving_method})   as follows   
\begin{align}\label{eq:l0}
\max_{\mat{h}} \  &\mat{D} \mat{h} + c_t ||\mat{L} * \mat{b}^T\otimes \mat{h}^T||_2  
 &\text{     s.t. }     	  h(n)\in[0,1], \ \forall n  
			 & ||\mat{h}||_0 \leq T_0
\end{align}
where we have used $\mat{X} = \mat{M}\mat{P}\mat{I}_{K+1} \otimes \mat{h}=    \mat{b}\otimes \mat{h}$.
The above problem maximizes a convex objective function over a polytope, defined by both constraints. If the objective function has a maximum value on the feasible region, then it is at the edges of the polytope. Therefore, the problem reduces to a finite computation of the objective function over the finite number of extreme points.

However, in the case of large networks, this computation could be too expensive. Therefore, we propose an algorithm that walks along the edges of the polytope. The intuition is similar to the one of the simplex algorithm or any hill climbing algorithm. 
Let consider an iterative algorithm, where at each iteration the $N$ variables $h_n$, with $n=1, \ldots, N$, are subdivided into basic variables and non-basic variables. The former are the ones such that $h_n=1$, while the non basic variables are the remaining zero sources. At each iteration, we perform  the operation of moving from a   feasible solution to an adjacent   feasible solution by swapping a basic variable with a non basic one (similar to the pivoting operation in the simplex algorithm). We move in such a way that the objective function always increases. We then stop the algorithm either after a maximum number of iteration steps or when convergence is reached. This is the \algolight, presented in  Algorithm \ref{algo_Iter} and further described in the following.
\begin{algorithm}[t] 
  \caption{\algolight} \label{algo_Iter}
  \begin{algorithmic}
\STATE     {\bf Input:}  
\STATE        number of iterations $MaxIter$, action sparsity level $T_0$, graph topology (and therefore $\mat{L}$ and $\mat{P}$), reward mask $\mat{M}$, estimated polynomial  $\hat{\mat{\alpha}}$, confidence bound $c$.      
\STATE     {\bf Output:}  
\STATE        optimal source signal $\mat{h}^{\star}$
\STATE     {\bf Initialization:}  
\STATE     Definition of the objective function $J(\mat{h})=\mat{M}\mat{P} \mat{I}_{K+1} \otimes \mat{h} \hat{\mat{\alpha}} + c ||\mat{L} * \mat{b}^T\otimes \mat{h}^T||_2$
\STATE      Evaluation of the partial derivatives  $a_n = \frac{\partial J}{\partial h_n} \bigr|_{\mat{u_n}}, \forall n$
\STATE      Selection of $\mat{h}^{(0)}$:  $h_{n}^{(0)}=1$ if $a_n$ belongs to the $T_0$ largest partial derivatives.  
\STATE      $t=1$
\FOR {$t\leq MaxIter$} 
	\STATE  Set $\mathcal{B}_{t}=\{ n |  h_{n}^{(t-1)} =1 \}$
	\STATE Evaluate the IN and OUT variables:  
         $$ {in} = \arg\max_{n | h_n \notin \mathcal{B}_{t}} \left\{    \frac{\partial J}{\partial h_n} \Bigr|_{\mat{h}^{(t-1)}}   \right\} , \ \ \ \ \ {out} = \arg\min_{n | h_n \in \mathcal{B}_{t}} \left\{    \frac{\partial J}{\partial h_n} \Bigr|_{\mat{h}^{(t-1)}}   \right\} $$
 	\STATE Set $\mat{h}^{(t)} = \mat{h}^{(t-1)}$
	\STATE Set  ${h}_{in}^{(t)}=1, {h}_{out}^{(t)}=0$ 
	\IF{$J(\mat{h}^{(t)}) \leq J(\mat{h}^{(t-1)})$}
	    \STATE $\mat{h}^{\star} = \mat{h}^{(t)}$
	    \STATE  {\bf break}
	    	 \ENDIF
        \STATE  $t \leftarrow t+1$
\ENDFOR 
\end{algorithmic}
\end{algorithm}

Let   $\mat{h}^{(t-1)}=[h_{1}^{(t-1)}, h_{2}^{(t-1)}, \ldots, h_{N}^{(t-1)}]$  be the optimal variable at the iteration step $i-1$. Let $\mathcal{B}_{t}=\{ n |  h_{n}^{(t-1)} =1 \}$ be the set of the indices of basic variables at $t$. Let then denote by $J$ the objective function $J(\mat{h})=\mat{M}\mat{P} \mat{I}_{K+1} \otimes \mat{h} \hat{\mat{\alpha}}_t  + c_t ||\mat{L} * \mat{b}^T\otimes \mat{h}^T||_2$ and by $\partial J/\partial h_n$ be the partial derivative of $J$ with respect the   $n$th variable. Finally, note  that vertices are adjacent if they share all but one non-basic variable. Equipped with the above notations and definitions, we now state the following Lemmas (proofs in Appendix \ref{sec:Appendix_algo_light}):

{\bf Lemma 3:}  \emph{One vertex is optimal if there is no better neighboring vertex.}

\vspace{.3cm}

{\bf Lemma 4:}  \emph{From a vertex,   moving to one of the neighboring nodes in the direction of the greatest gradient leads to a no-worse objective function. Let $h_{in}^{(t)}$  and $h_{out}^{(t)}$  be the variable that enters and leaves the set of basic variables, respectively, at the $t$-th iteration. These variables are evaluated as follows }
   $$ {in} = \arg\max_{n | h_n \notin \mathcal{B}_{t}} \left\{    \frac{\partial J}{\partial h_n} \Bigr|_{\mat{h}^{(t-1)}}   \right\}, \ \ \ \ 
    {out} = \arg\min_{n | h_n \in \mathcal{B}_{t}} \left\{    \frac{\partial J}{\partial h_n} \Bigr|_{\mat{h}^{(t-1)}}   \right\} $$
\newline
\
\newline
From Lemma 4, given a vertex  $\mat{h}^{(t-1)}$, at the $t$-th iteration the algorithm will move to the neighboring vertex  $\mat{h}^{(t)}$ defined as follows: 
 $$h_{\text{in}}^{(t)}=1,\ \  h_{\text{out}}^{(t)}=0,  \ \ \text{  and   }  \ \ h_{n}^{(t)}=h_{n}^{(t-1)}, \forall n\neq in, out.$$
As shown in Algorithm \ref{algo_Iter},  if the swap variable leads to an improvement of the objective function, \emph{i.e.}, if $J(\mat{h}^{(t)}) > J(\mat{h}^{(t-1)})$, then we proceed to the next step. Otherwise, we set the optimal source signal $\mat{h}^{\star} = \mat{h}^{(t-1)}$ and we break the iterative loop. Further details are provided in Algorithm \ref{algo_Iter}, together with  the initialization step. Rather than a randomly generating starting point, we consider the one with the $T_0$ variables having the maximum partial derivative $a_n = \frac{\partial J}{\partial h_n} \bigr|_{\mat{u_n}}, \forall n$, with  $\mat{u_n}$ being a $N$-dimensional vector all elements null but the  $n$-th, which is set to 1. Note also that in the algorithm the  partial derivative of the objective function  can be derived as (see Appendix~\ref{sec:Appendix_algo_light_derivative} for details) 
\begin{align}
\frac{\partial  J(\mat{h})}{\partial h_n} &=  \mat{M}\mat{P}    \mat{I}_{K+1} \otimes \mat{1}_n \hat{\mat{\alpha}} +   \frac{  \left(\mat{L} * \mat{b}^T\otimes \mat{h}^T \right) }{ ||\mat{L} * \mat{b}^T\otimes \mat{h}^T||_2}   \left(\mat{L} * \mat{b}^T\otimes \mat{1}_n^T \right)^T
\end{align}

In summary, the proposed algorithm requires the evaluation of the partial derivative ($N$ operations) instead of exhaustively evaluating the objective function in \eqref{eq:l0} at all  ${ N \choose T_0}$  possible edges.  In the following Section, we show emporically that this   complexity reduction does not come at the price of reduced optimality.  

\section{Simulation Results}
\label{sec:results} 

\subsection{Settings}
As benchmark solution, we propose an algorithm denoted as Act After Learning (AAL), in which the exploration and the exploitation phases are separated, while our proposed method finds the best tradeoff between exploitation and exploration automatically. 
The key intuition is that it first gathers a training set (in the first $T_L$ decision strategies) and therefore experience a reward as a function of random actions. Then, after a training phase of $T_L$ decision opportunities, the generating kernel is estimated and the best arm is selected. In the remaining decision opportunities the best action is taken. 
 Note that we do not compare with the linUCB algorithms since its regret would scale with the cardinality of the decision space   $|\mathcal{A}|={N \choose T_o}$, if $T_0$ is the imposed sparsity of $\mat{h}$, which is prohibitive in the case of large-scale network. .

We carry out experiments on Barab{\'a}si-Albert model (BA)  graphs \cite{BarabAsi509},  on  radial basis function (RBF) random graphs, and on non-synthetic graphs (\emph{e.g.}, Minnesota graph\footnote{Available at \texttt{https://lts2.epfl.ch/gsp/}.}). For BA graphs, the network begins with an initial connected network of  $m_{0}=10$ nodes. At each iteration, one node is added to the network and it is connected to   $m\leq m_{0}$ existing nodes. Connections to existing nodes are built following a preferential attachment mechanism, which eventually builds a scale-free network. For the RBF model, we generate the coordinates of the vertices uniformly at random in the unit square, and we set the edge weights based on a thresholded Gaussian kernel function so that $W(i,j) = \exp(-[dist(i,j)^2)/2\sigma)$  if the distance between vertices $i$ and $j$ is smaller than or equal to $T$, and zero otherwise. We further set $\sigma = 0.5$ and we vary $T$ to change the edge density of the generated graphs.   

We model the network processes as a diffusion processes  with generating kernel $g_L = e^{-\tau L}$~\cite{Thanou:J17}. We then consider that each signal on the graph is characterized by the source signal, the generating kernel and an additive random noise $\epsilon_t$ with zero mean and variance $\sigma_e^2$ (i.e., $R=\sigma_e$ in the spectral UCB).  The remaining parameters of the sequential decision strategy are set as $\mu=0.01$, $\delta=0.01$. The mask $M$ is randomly generated and it covers $20\%$ of the nodes. 

\subsection{Performance of \algo}
\begin{figure*}[t]
\centering 
\subfigure[Random Graph, Polynomial Dictionary ($K=20$)]{\includegraphics[width=.4\linewidth,  draft=false]
{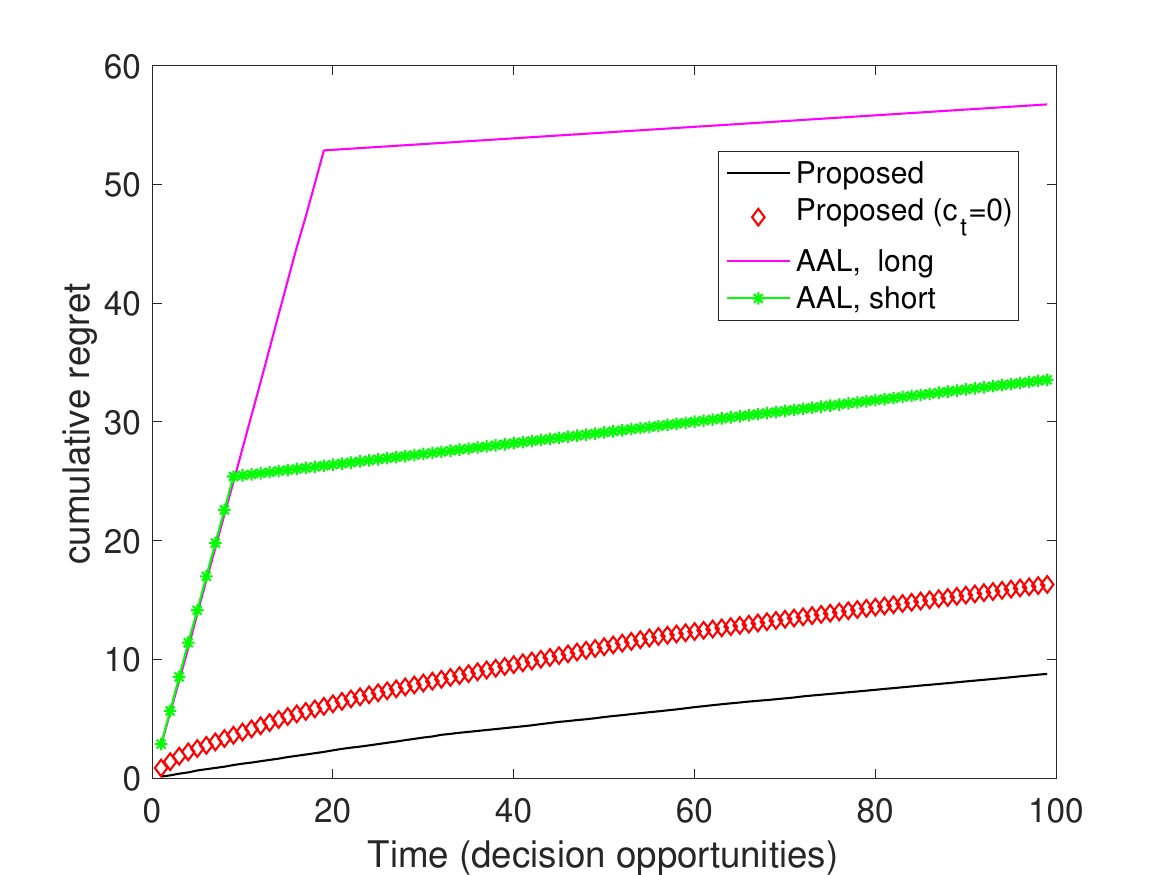}\label{fig:RandG_PolyDict_localMASK_T05_noise01_m01_N100_mask05_Diff0_tau20_K20}} \hfil
\subfigure[Community Graph, Diffusion dictionary, $\tau=5$. ] { \includegraphics[width=.4\linewidth,  draft=false]{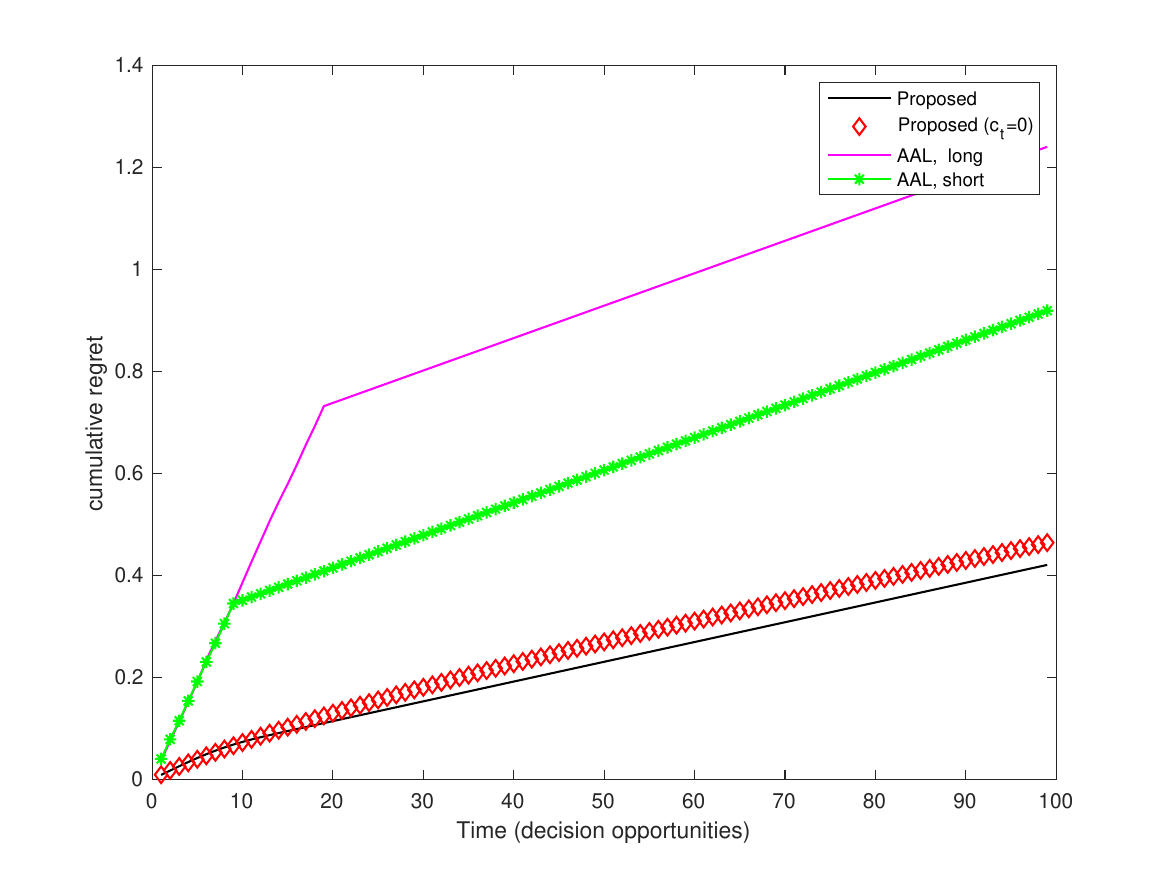}
\label{fig:Community_local1_T05_noise0.01_m01_N256_mask05_Diff1_tau5_K20}
}  \caption{Cumulative regret vs. time for randomly generated graphs with $N=100$, diffusion process (with $\tau=5$) and sparsity level $T_0=5$ for   \algo.} \label{fig:LoadedG_Diff_T02_Noise2}
\end{figure*}

We now  study the performance of the proposed \algo \,  with \algolight \, used  in Step 2. First, a randomly generated graph (RBF model) with $N=100$ nodes is considered, in the case of diffusion process acting on the graph with $\tau=10$ and with $\sigma_e^2=10^{-2}$.  Fig. \ref{fig:LoadedG_Diff_T02_Noise2} depicts the cumulative regret  over time  (in terms of decision opportunities) for the considered graph. Each point is averaged over $100$ realizations (when at each realization both the graph and the noise of the signal on graph are generated).  The \algo \, is compared to AAL with   $T_L=10$ (AAL short) and $T_L=20$ (AAL long). Note that a longer exploration leads to a better estimate but for a longer (suboptimal) exploration phse.   From Fig.  \ref{fig:LoadedG_Diff_T02_Noise2}, we observe that \algo \, outperforms the baseline algorithms in both networks. The proposed algorithm is tested also in the case of $c_t=0$, which means that the confidence bound is ignored when acting -- leading to less exploration. The comparison shows the gain in striking the optimal tradeoff between exploitation and eploraztion ($c_t>0$). 
Further results   are provided in Appendix \ref{subsection:comparison}.

\begin{figure*}[t]
\centering 
\subfigure[Source signal (red) and mask signal $M$ (blue)]{\includegraphics[width=.33\linewidth,  draft=false]{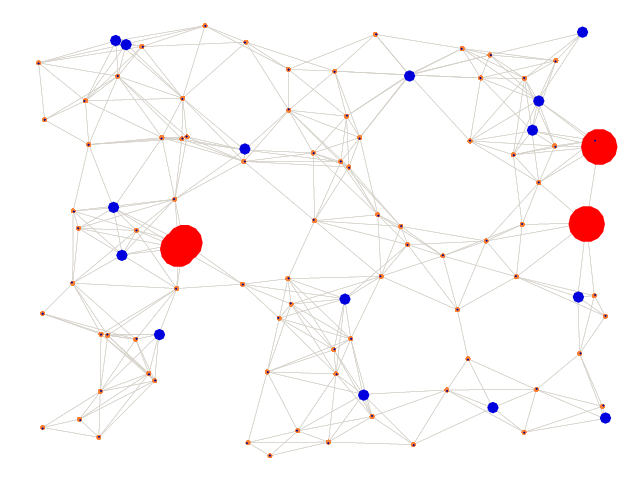}\label{fig:Rndm_N100_T04_Tau4_H}} \hfil
\vspace{0.1cm}
\subfigure[Resulting Signal] { \includegraphics[width=.33\linewidth,  draft=false]{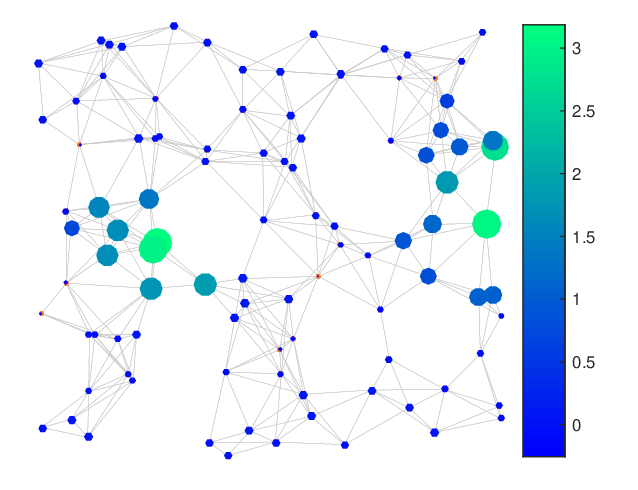}
\label{fig:Rndm_N100_T04_Tau4}} 
\caption{Optimal source signal and resulting signal for a randomly generated graph (RBF model) with $N=100$, and sparsity level $T_0=4$.} \label{fig:Rndm_N100_T04_Tau4_H_Signal}
\end{figure*}

We further illustrate in Fig. \ref{fig:Rndm_N100_T04_Tau4_H_Signal} that optimal placement of sparse resources in high dimensional networks is not necessarily an intuitive step. It depicts the optimal source signal computed by \algo \ with optimal solver introduced in Appendix \ref{appendix:solving_method} and the resulting signal for a randomly generated graph (RBF model) with $N=100$, and sparsity level $T_0=4$. In Fig. \ref{fig:Rndm_N100_T04_Tau4_H}, the optimal signal is depicted in red, while the mask signal used to evaluate the reward is depicted in blue.  Interestingly, the optimal signal is placed on nodes that do not necessarily belong to the mask and do not necessarily appear to be central in the graph. Yet, this results in the optimal reward signal depicted in Fig. \ref{fig:Rndm_N100_T04_Tau4}.

\begin{figure*}[t]
\centering 
\subfigure[CPU Time]{\includegraphics[width=.35\linewidth,  draft=false]{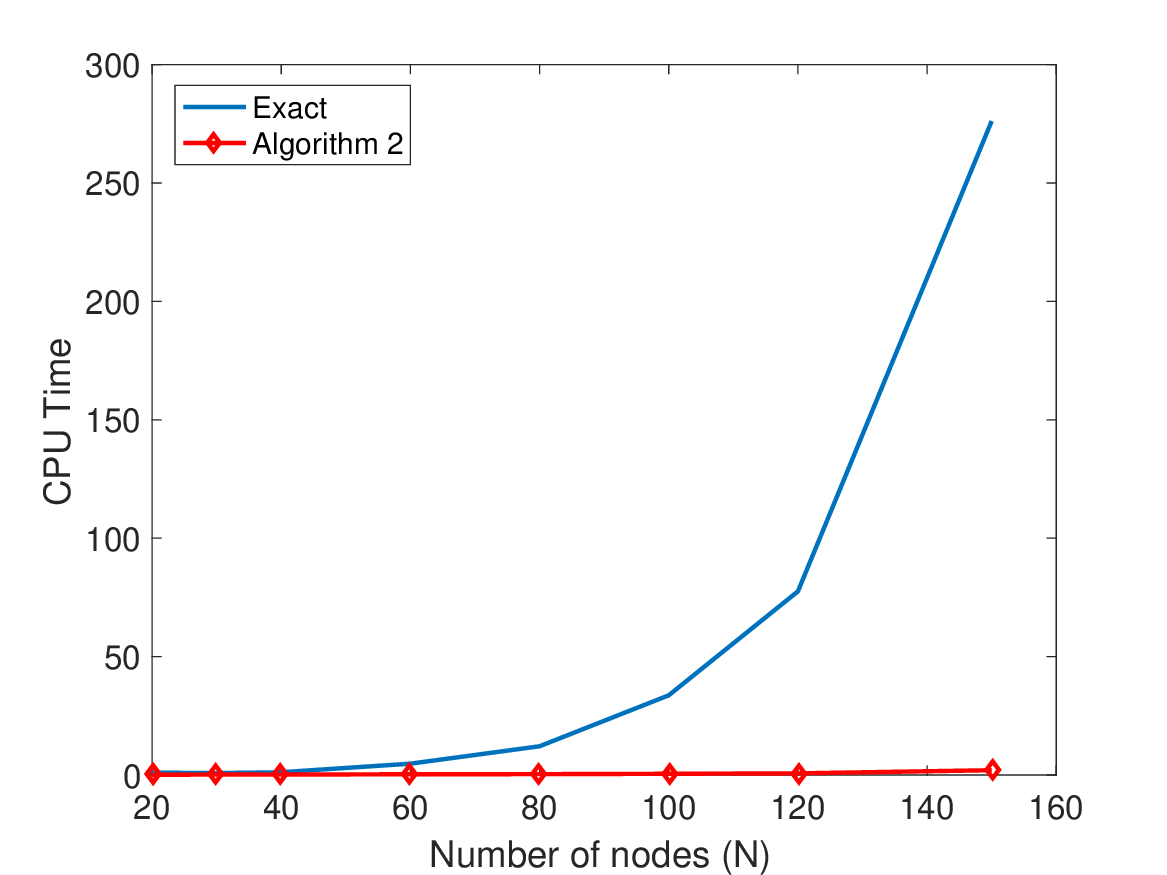}\label{fig:CPUTime_Vs_N}} \hfil
\vspace{0.1cm}
\subfigure[Reward] { \includegraphics[width=.35\linewidth,  draft=false]{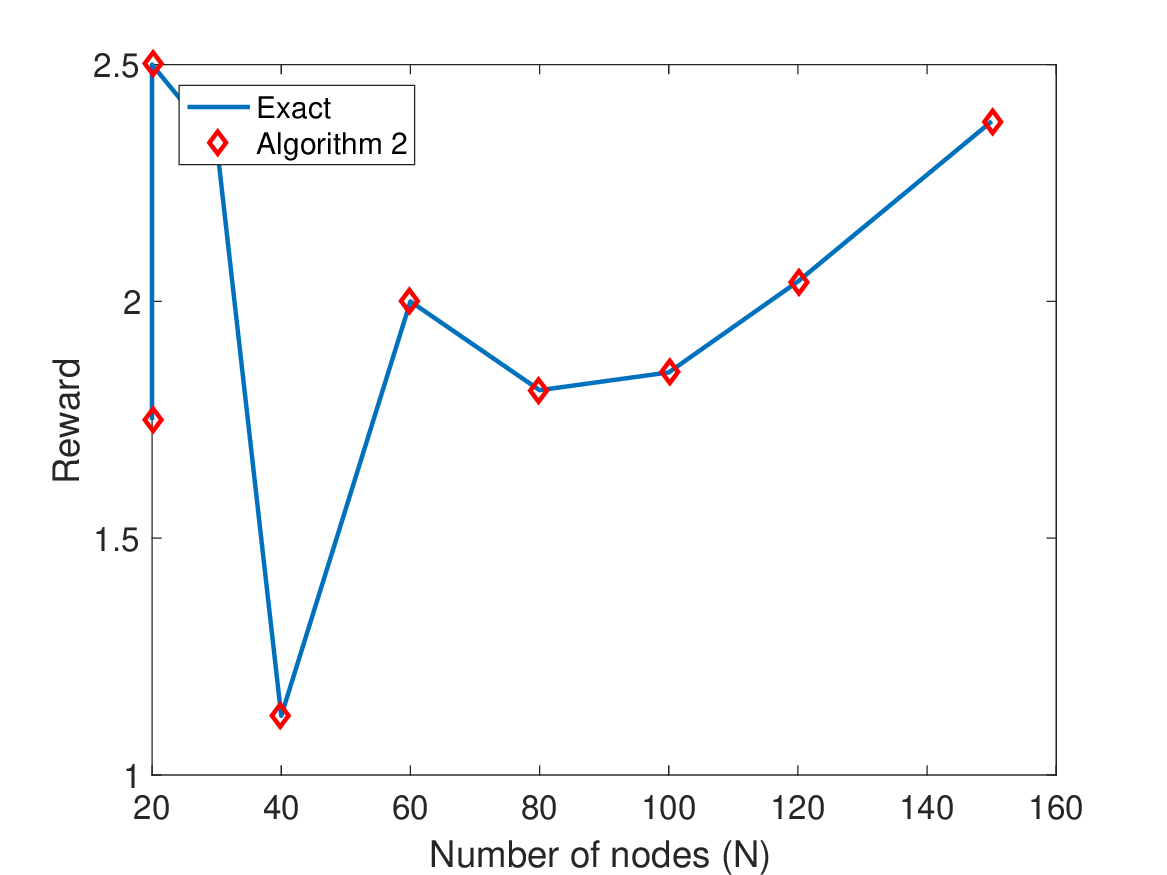}
\label{fig:Reward_vs_N}
} 
\caption{Comparison of the optimal solver and \algolight (Algorithm 2) for random graphs (RBF model) with different number of nodes, and $T_0=5$.} \label{fig:complexity_vs_n}
\end{figure*}

We validate now the proposed algorithm for solving the action selection step, which is a priori NP-hard. We empirically compare the numerical solver (FMICON) adopted to optimally solve Step 2 (labelled in the following figure as ``Exact") and the \algolight \, (labelled as ``Algorithm 2"). Fig. \ref{fig:CPUTime_Vs_N} depicts the CPU time required by both solvers as a function of the number of nodes $N$ for a randomly generated graph (RBF model), with sparsity level $T_0=5$. The achieved reward after $100$ decision steps is also depicted in  Fig. \ref{fig:Reward_vs_N}. Results are averaged over $50$ generated graphs. 
As expected, the problem in \eqref{eq:l0} is NP-hard (maximization of a convex function under convex -or affine- constraints) and the solver's complexity grows exponentially with $N$. Conversely, the complexity of \algo \, grows linearly with $N$, as shown by Fig.  \ref{fig:CPUTime_Vs_N}.  From Fig. \ref{fig:Reward_vs_N}, it can be observed that the proposed solution still achieves the optimal solution in terms of reward. Note that the reward is not monotonic with $N$ because the density of the graph is not necessarily kept constant for different $N$ values. 

\begin{figure*}[t]
\centering
\subfigure[Graph Laplacian Eigenvalues]{\includegraphics[width=.4\linewidth,  draft=false]{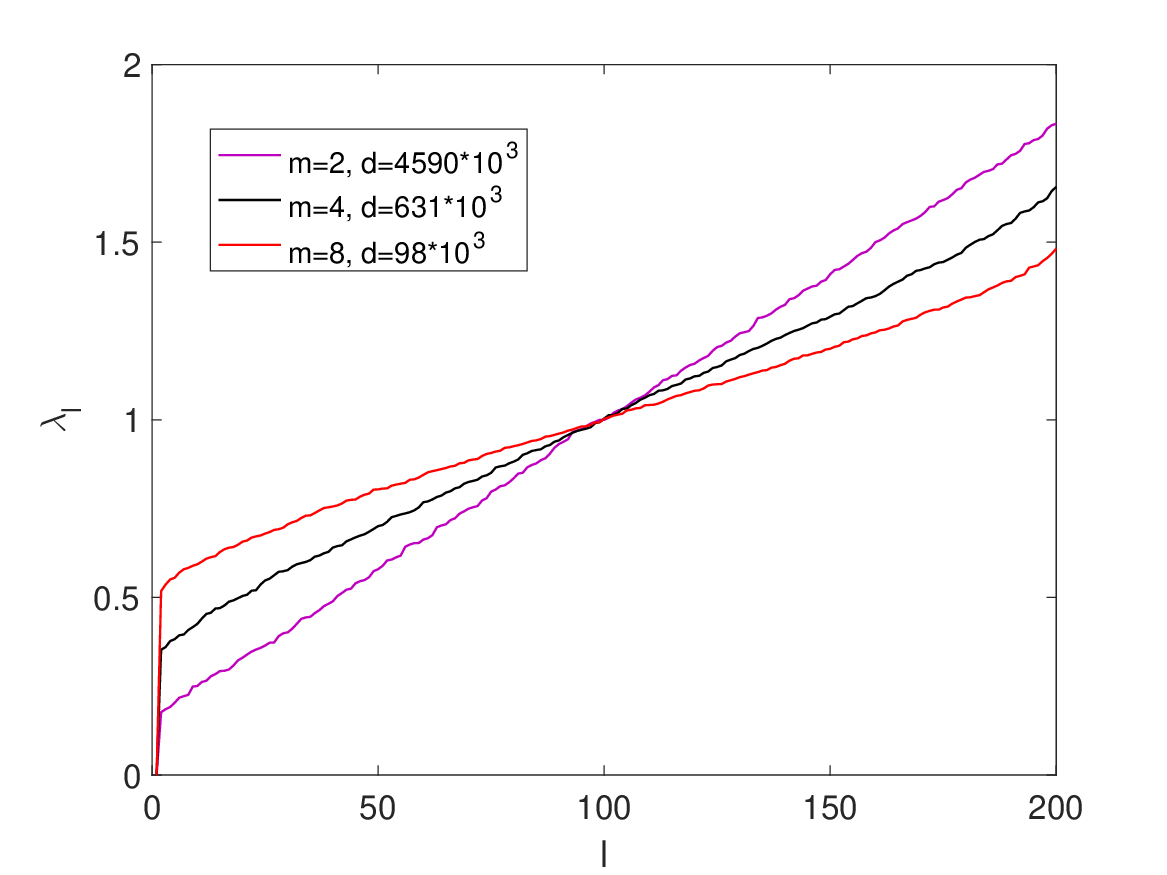}\label{fig:Laplacian_Comparison_N300_BA_2}} \hfil
\vspace{0.1cm}
\subfigure[Estimation Error] { \includegraphics[width=.4\linewidth,  draft=false]{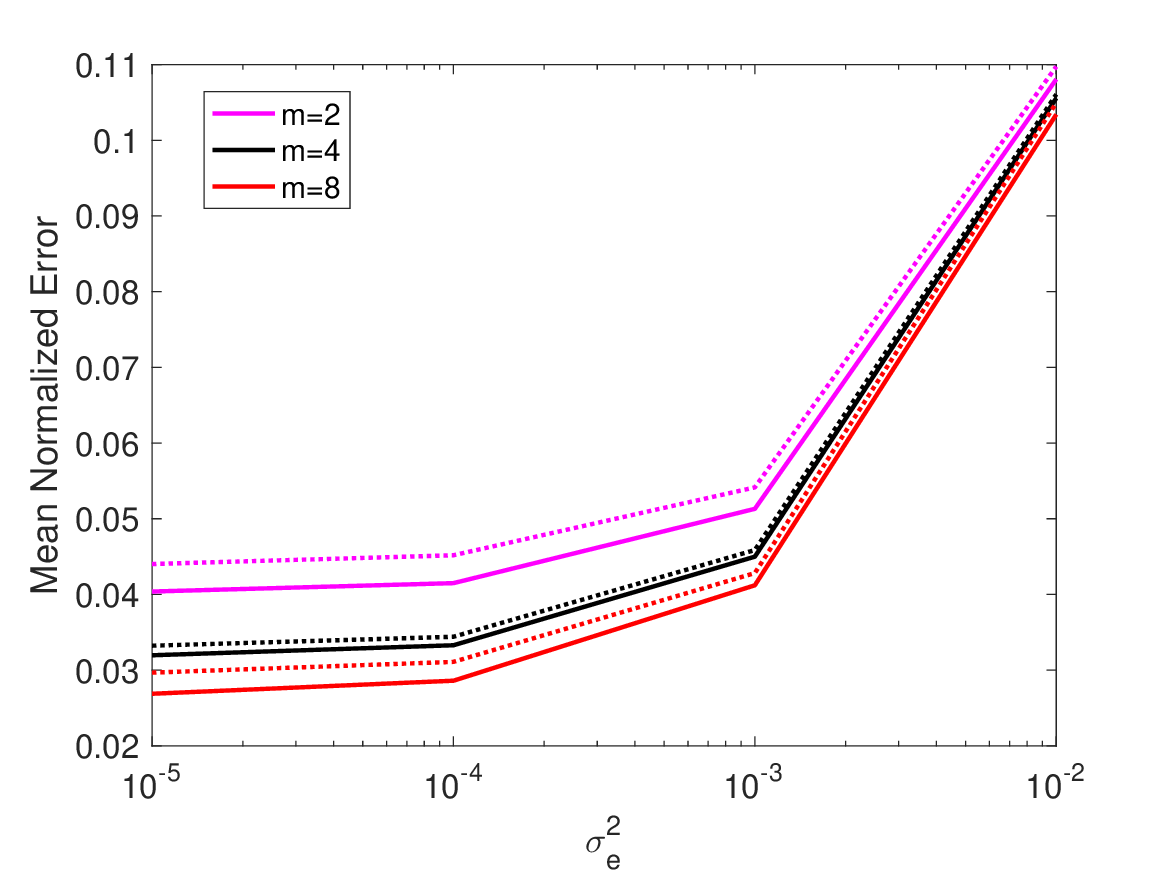} \label{fig:Error_Vs_m0_BA_200_Mask_A}} 
\caption{Graph Laplacian distribution and signal estimation error for  graphs generated with the BA model with different levels of connectivity,  $N=200$ nodes, and sparsity value $T_0=25$. The estimation error is evaluated both in the case of full observability (solid line) or in the case of partial observability (dotted line). } \label{fig:laplacian_error_BA}
\end{figure*}

Finally, we evaluate the effect of network topology on the estimation error  (From Lemma 2,  we see that  the confidence bound increases with  the sparsity level $T_0$ and $d  =\sum_{k=0}^K \sum_{l=1}^N \lambda_l^k $). 
We    consider a randomly generated training set of $300$ signals, and we then estimate the accuracy of the learned polynomial $\alpha$.   To measure the accuracy of the estimate, we evaluate the error on the resulting signal given the action $\mat{h}$ of test signals.  
We consider graphs generated with the BA model; by changing the  parameter $m$, we generate more or less connected graphs (the larger the $m$ the more connected is the graph). This is reflected in the power sum $d$ and in a more narrow profile of the eigenvalues of the Laplacian $\lambda_l$, as observed from Fig. \ref{fig:Laplacian_Comparison_N300_BA_2}, where the values of  $\lambda_l$ are provided for different graph topologies.  As a consequence, more connected graphs lead to a more accurate estimate of the generating kernels, see Fig. \ref{fig:Error_Vs_m0_BA_200_Mask_A}. This validates the understanding we had from the theoretical bounds, which depend on $d$.

\section{Related Work}
\label{sec:related works}
Multiple works have looked at graph-based bandit problems with the ultimate goal of addressing sample-efficiency~\cite{slivkins2014contextual,valko2015cheap,Bellemare:A19,Mohaghegh:J19,waradp2020deep,ide2022targeted},   identifying and leveraging the structure underneath data in optimisation  problems where the outputs have semantically rich structure. In this direction, graph knowledge has been used for 1)    sharing the payoff throughout the graph Laplacian~\cite{pmlr-v117-lykouris20a,lee2020closer,ghari2022online,esposito2022learning} or reducing the dimensionality of the search space by clustering arms~\cite{gentile2014online,li2015data, cesa2013gang,LiGKZ16,li2016collaborative,korda2016distributed,Caron:A12}   --  those usually perform poorly in irregular datasets~\cite{Yang:A18}, typical of most of the real-world problems;
2) 
modelling each arm as graph~\cite{kassraie2022graph} -- the main goal is optimization over graph domains (set of graphs) while our focus is optimizing over geometrical signal domains (set of signals on a given graph); 3) modelling the reward signal, seen as smooth signal on the graph~\cite{valko2014spectral}-- this shares most similarities with our work hence they are further describe in the following paragraph where differences with respect to our works are also highligthed.

Looking at the arms as nodes on a graph and the reward as a smooth signal on this graph~\cite{valko2014spectral,kocak2014spectral,hanawal2015,valko2015cheap} permits to i) define the reward as a linear combination of the eigenvectors of the graph Laplacian matrix, where the linear coefficients are unknown, ii) apply  LinUCB \cite{chu2011contextual} in the spectral domain.   These algorithms achieve a regret bound of the order $\sqrt{dT}$, with $d$ being the effective dimension (linked to the dimensionality of the characteristic eigenvalues) and $T$ being the number of rounds. Similar intuitions have been introduced in \cite{valko2013finite}, which performs  maximization over the smooth functions that have a small \ac{RKHS} norm,  or in \cite{kaige_aistat20,thaker2022maximizing} that exploit graph homophily   to denoise/generalize the reward.
Similarly, our work exploits spectral graph prior to  solve a linUCB algorithm with the assumption of  low dimensional reward behavior. Yet, we expand the literature on spectral MABs by modelling processing evolving on graphs. Namely, each action represents possibly a set on a graph (not limited to a node only) and the reward is not  necessarily smooth on the graph and it is a resultant signal on the entire graph instead of on one node only. 

Our work, as well as spectral MAB ones, can be seen as extensions of the rich field of methods for kernelized bandits working under the norm bounded \ac{RKHS} assumption~\cite{valko2013finite,pmlr-v70-chowdhury17a,pmlr-v139-camilleri21a,pmlr-v119-yang20h} but they are not limited to Euclidean domains. Another line of works consider bandits theory for decision-making strategy~\cite{pmlr-v119-yang20h, hsieh2023thompson} (and many references therein), which however differ from our model in which we limit decisions to bandit problems and we exploit GSP tools to learn the dynamic process of the network flow.  Our paper share the ultimate goal of~\cite{ide2022targeted}, focused on online targeted advertising on social networks, where the  multi-mode tensor tool can nicely complement our work to extend our signal processing analysis to heterogeneous feature vectors. 
 Complementary to this work, there is also the vast literature on Thompson sampling for the  multi-arms bandit problem that could be an interesting alternative to our  \algolight  \cite{agrawal2012analysis} 

Finally, there is a vast literature from the GSP community 
~\cite{ortega2018graph} aimed at capturing structural properties of network processes (i.e., node centrality~\cite{he2021detecting}, community detection~\cite{wai2019blind}) for network problems such as diffusion dynamics, pricing experiments, and opinion dynamics: the work in \cite{wai2019blind} models an unknown network process as a graph filter that is excited by a set of unknown low-rank inputs, the study in\cite{Raksha:J21} models power systems as generative low-pass graph filters. 
However,
no work so far has focused on learning while acting, i.e., inferring network process models while taking sequential actions on those networks.

\section{Conclusions}
\label{sec:conclusion}

In  this  work, we  study network optimization problems under uncertainty in the case of optimal source placement.
As main contributions, we cast the network optimization problem under uncertainty as a linear MAB problem, which infers a $K$-dimensional polynomial that defines the graph generating-kernel while taking actions over time on the network-graph.  We  then derive the theoretical bound of the estimation of the graph spectral model and translate it  to  the  MAB  upper  confidence  bound.  We  show  both  mathematically and  empirically  that  more  connected  graphs  and  sparser  signals  lead  to  a  more  accurate estimation of the network processes. Finally,  we  observe  that  the  optimization  method  leads  to  an  arm  selection  problem  that is NP-hard, and we provide a low-complexity algorithm by exploiting the structure of the optimization  function.  Beyond proposing a data-efficient solution to problems of network optimization, this work aims at opening the gate to new research directions in which graph signal processing tools are blended to online learning frameworks to exploit structural knowledge of network optimization problems. 


\newpage

\section{Ethical Statement}


Our work is mostly of theoretical nature, and we do not foresee any direct ethical implications. There is always a risk, as for most works of theoretical and algorithmic nature in machine learning, that the work would be diverted from its original objective, and largely modified to design extensions in non-ethical applications. However, this is not obviously envisaged by the authors at the time of the writing.

\bibliographystyle{splncs04}
\bibliography{ML_GRAPH,GSP} 

\clearpage

 \setcounter{page}{1} 
 
\appendix

\section{Proof of Lemmas}
\label{sec:appendixA}

\emph{{\bf Lemma 1:} Suppose $\mat{Z}_1,  \mat{Z}_2, \ldots,  \mat{Z}_t \in \mathbb{R}^{1\times (K)}$, with $\mat{Z}_{\tau}=\mat{M}\mat{P}\mat{I}_{K} \otimes \mat{h}_{\tau}$ and for any $1\leq \tau \leq t-1$, $||h_{\tau}||_F^2\leq T_0$, and $||\mat{M}||_F^2\leq Q$. Let $\mat{V}_t = \sum_{\tau} \mat{Z}_{\tau}^T \mat{Z}_{\tau} + \mu \mat{I}_{K}$ with $\mu>0$,  then}
$
|\mat{V}_t| \leq \left[ \mu +   d QT_0 \right]^{K},
$
\emph{with $d=\sum_k \sum_l \lambda_l^k$, with $\lambda_l$ being the $l$-th eigenvalue of the graph Laplacian. }

\

\emph{Proof}:  
As shown in \cite{Abassi} the following inequality holds $|\mat{V}_t| \leq (Tr(\mat{V}_t)/K)^{K}$. We now look at the trace of $\mat{V}_t$:
\begin{align}
Tr(\mat{V}_t) = Tr(\mu \mat{I}_{K}) + Tr\left(\sum_{\tau=1}^t \mat{Z}_{\tau}^T \mat{Z}_{\tau}\right) = K\mu +  \sum_{ \tau =1}^t Tr( \mat{Z}_{\tau}^T \mat{Z}_{\tau})
\end{align}
\begin{align}
Tr( \mat{Z}_{\tau}^T \mat{Z}_{\tau}) &= ||\mat{Z}_{\tau}||_F^2 
= || \mat{M}  \mat{PH}||_F^2 \leq  Q \,  ||\mat{P}||_F^2  \,  ||\mat{I}_{K}||_F^2  \,  ||\mat{h}_{\tau}||_F^2  \nonumber \\
&= KQ ||\mat{P}||_F^2  \,    ||\mat{h}_{\tau}||_F^2  \leq  KQ  T_0 ||\mat{P}||_F^2
\end{align}
 where the first inequality comes from $||\mat{M}||_F^2     \leq Q$, and  the last inequality comes from the imposed sparsity level on $\mat{h}$ and from the condition $|h_n|\leq1$.
Finally, we look at the Frobenius norm of $\mat{P}$ as follows
\begin{align}
||\mat{P}||_F^2 = \sum_k ||\Lambda^k||_F^2 = \sum_k \sum_l \lambda_l^k = d
\end{align}
where $\Lambda$ is the diagonal matrix with entries the graph Laplacian eigenvalues $\lambda_l$. 
Thus, we can bound the trace of $\mat{V}_t$ as
\begin{align}
Tr(\mat{V}_t)   \leq  K\mu +  KQ  T_0 d
\end{align}
and then derive the inequality of Lemma 1. $\square$

\

\emph{{\bf Lemma 2:} Assume that $\mat{V}_t = \sum_{\tau} \mat{Z}_{\tau}^T \mat{Z}_{\tau} + \mu \mat{I}_{K}$, define ${\mat{w}}_{\tau} = \mat{Z}_{\tau} \mat{\alpha_*} + \mat{\eta}_{\tau}$, with $\mat{Z}_{\tau}=\mat{M}\mat{P}\mat{I}_{K} \otimes \mat{h}_{\tau}$  and with $\mat{\eta}_t$ being conditionally $R$-sub-Gaussian, and assume that $||\mat{\alpha}_*||_2\leq S$, and $||\mat{h}_{\tau}||_F^2\leq T_0$. Then, for any $\delta>0$,  with probability at least $1-\delta$, for all $t\leq0$, $\mat{\alpha}_*$ lies in the set }
 $$
E_t: \left\{ {\mat{\alpha}} \in \mathbb{R}^{1\times K} :   || \hat{\mat{\alpha}}_t - \mat{\alpha}||_{\mat{V}_t}  \leq   R\left[ \sqrt{K \log(\mu+tdQT_0 ) +  2 \log (\mu^{-1/2}\delta) } \right] + \mu^{1/2}S  
\right\}
$$
\emph{with $d=\sum_k \sum_l \lambda_l^k$, with $\lambda_l$ being the $l$-th eigenvalue of the graph Laplacian and $\hat{\mat{\alpha}}_t$ is the $l^2$-regularized least-square estimate of $\mat{\alpha}_*$ when $t$ training samples are available. }

\

\emph{Proof:} We now study the LOWER bound on the estimated  $\hat{\mat{\alpha}}_t$, recalling that the vector is estimated observing the masked signal  ${\mat{w}}_t$.  Similarly to \cite{Abassi}, using
\begin{align} 
\hat{\mat{\alpha}}_t  &=  \mat{V}_t^{-1} \mat{Z}_{1:t}^T \mat{\tilde{Y}}_{1:t}^T    =  \mat{V}_t^{-1} \mat{Z}_{1:t}^T \left(\mat{Z}_{1:t} \mat{\alpha}^{\star}  + \mat{\eta}_{1:t} \right)  \nonumber \\
  &=\mat{V}_t^{-1} \mat{Z}_{1:t}^T   \mat{\eta}_t    + \mat{V}_t^{-1} \mat{Z}_{1:t}^T \mat{Z}_{1:t} \mat{\alpha}^{\star} + 
\mu \mat{V}_t^{-1}  \mat{\alpha}^{\star} -  \mu \mat{V}_t^{-1}  \mat{\alpha}^{\star} \nonumber \\
  &= \mat{V}_t^{-1} \mat{Z}_{1:t}^T   \mat{\eta}_{1:t}   + \mat{V}_t^{-1} \left( \mat{Z}_{1:t}^T \mat{Z}_{1:t}  + \mu \mat{I}_{K+1} \right)   \mat{\alpha}^{\star} -  \mu \mat{V}_t^{-1} \mat{\alpha}^{\star} \nonumber \\
  &= \mat{V}_t^{-1} \mat{Z}_{1:t}^T   \mat{\eta}_{1:t}    +     \mat{\alpha}^{\star} -  \mu \mat{V}_t^{-1}  \mat{\alpha}^{\star}\,.
\end{align}   
Let's now consider a vector $x\in\mathbb{R}^{K\times 1}$, we get 
\begin{align} 
\mat{x}^T \hat{\mat{\alpha}}_t - \mat{x}^T  \mat{\alpha}^{\star}  &=  \mat{x}  \mat{V}_t^{-1} \mat{Z}_{1:t}^T   \mat{\eta}_{1:t} -   \mu \mat{x}\mat{V}_t^{-1} \mat{\alpha}^{\star} = \langle\mat{x}, \mat{Z}_{1:t}^T   \mat{\eta}_{1:t}\rangle_{V_t^{-1}} -\mu  \langle\mat{x}, \mat{\alpha}_*\rangle_{V_t^{-1}}
\end{align}   
 Using Cauchy-Schwarz, we get 
\begin{align} 
|\mat{x}^T \hat{\mat{\alpha}}_t - \mat{x}^T \mat{\alpha}_*|  &\leq  ||\mat{x}||_{V_t^{-1}} \left( || \mat{Z}_{1:t}^T   \mat{\eta}_{1:t}||_{V_t^{-1}} + \mu ||\mat{\alpha}_*||_{V_t^{-1}}   \right)  \nonumber \\
&\leq   ||\mat{x}||_{V_t^{-1}}   R\sqrt{2 \log \frac{|\mat{V}_t|^{1/2} }{\delta  |\mu \mat{I}_{K}|^{1/2}} }  + \mu^{1/2}S 
\end{align}   
where the last inequality comes from Theorem 1 in \cite{Abassi}, under the condition that $\eta_t$ is conditionally R-sub-Gaussian  for some $R\geq 0$, and that $||\mat{\alpha}_*||_2\leq S$. In our case, $\mat{\eta}\sim\mathcal{N}(0,N\sigma_e^2)$, which means that it is a R-subgaussian variable with $R=\sqrt{N}\sigma_e$. 

Using $\mat{x}=\mat{V}_t (\mat{\alpha}_t - \mat{\alpha}^{\star})$ and $||\mat{V}_t (\mat{\alpha}_t - \mat{\alpha}^{\star})||_{\mat{V}_t^{-1}} = || \mat{\alpha}_t - \mat{\alpha}^{\star} ||_{\mat{V}_t}$, we get    
\begin{align} 
|| \mat{\alpha}_t - \mat{\alpha}^{\star} ||^2_{\mat{V}_t} &\leq     ||\mat{V}_t (\mat{\alpha}_t - \mat{\alpha}^{\star})||_{V_t^{-1}}  \left(   R\sqrt{2 \log \frac{|\mat{V}_t|^{1/2} }{\delta   |\mu \mat{I}_{K+1}|^{1/2}} }  + \mu^{1/2}S \right).
\end{align}   
 Diving both sides by $|| \mat{\alpha}_t - \mat{\alpha}^{\star} ||$ we obtain
 \begin{align} 
|| \mat{\alpha}_t - \mat{\alpha}^{\star} ||_{\mat{V}_t} &\leq   R\sqrt{2 \log \frac{|\mat{V}_t|^{1/2} }{\delta  |\mu \mat{I}_{K+1}|^{1/2}} }  + \mu^{1/2}S 
\end{align}   
Applying Lemma 1, we obtain
 \begin{align} 
 || \hat{\mat{\alpha}}_t - \mat{\alpha}||_{\mat{V}_t}  \leq   R\left[ \sqrt{K \log(\mu+tdQT_0 ) +  2 \log (\mu^{-1/2}\delta) } \right] + \mu^{1/2}S  \end{align}   
 This proves the confidence bound defined in Lemma 2. $\square$

\section{UCB and Regret Bound Derivation}
\label{sec:bound_variance}
\subsection{UCB Derivation}
By least square regression, we estimate the  mean reward, but we can also estimate   the variance of   reward, denoted by $\sigma_{\alpha}^2$,  
i.e. the uncertainty due to parameter estimation error. We can then define the UCB to be $c$ standard deviations above the mean by adding on a bonus for uncertainty,  $c\sigma_{\alpha}$ to the mean reward.  This leads to maximize the  
reward summed on the UCB. 

Being a linear regression, the parameter covariance is $V_t^{-1}$. being the reward linear in features $(\mat{X}\mat{\alpha})$, we obtain a quadratic reward variance $\mat{X}^T V_t^{-1} \mat{X}$.  The geometric interpretation is that we maximize the reward for any parameter vector $\alpha$ within an ellipsoid $E_t$ defined by $c$. It follows 
\begin{align}
\mat{h}_t &= \arg\max_{\mat{h}\in{\mathcal{A}}} \max_{\alpha\in E_t}   \mat{X \alpha}    \nonumber \\
&=  \arg\max_{\mat{h}\in{\mathcal{A}}} \left(  \mat{X \alpha} +    c_t \sigma_{X\alpha} \right) \nonumber \\
&=  \arg\max_{\mat{h}\in{\mathcal{A}}} \left(  \mat{X \alpha} +    c_t\sqrt{\mat{X}\mat{V}_t^{-1}\mat{X}^T}  \right) \nonumber \\
&= \arg\max_{\mat{h}\in{\mathcal{A}}}    \mat{X \alpha} + c_t ||\mat{X}||_{\mat{V}_t^{-1}} \nonumber \\
&= \arg\max_{\mat{h}\in{\mathcal{A}}}   \left[\mat{M}\mat{PH}\hat{\mat{\alpha}}_t  + c_t || \mat{M}\mat{PH}||_{\mat{V}_t^{-1}}\right]\; 
\end{align}
with  $c_t = R\left[ \sqrt{K \log(\mu+tdQT_0 ) +  2 \log (\mu^{-1/2}\delta) } \right] + \mu^{1/2}S $ following Lemma 2. 

\subsection{Regret Bound Derivation}
 We now derive the regret bound of \algo. From \cite{valko2014spectral}, we have
\begin{align}\label{eq:Regret_1}
R_T & \leq 2(c_T +1) \sqrt{T\sum_{t=1}^T \min\left( 1, ||\mat{X}||_{V_t^{-1}}^2  \right)} 
\end{align}
and from Lemma 11 in \cite{Abassi}, we also have 
\begin{align} \label{eq:Regret_2}
 \sum_{t=1}^T \min\left( 1, ||\mat{X}||_{V_t^{-1}}^2  \right)  \leq 2\log(|\mat{V}_t|/ \mu |\mat{I}_{K+1}|) = 2\log(|\mat{V}_t|/ \mu)
\end{align}
Substituting \eqref{eq:Regret_2} in \eqref{eq:Regret_1} and applying Lemma 1, we get 
\begin{align}
R_T & \leq 2(c_T +1) \sqrt{  2KT \log \left( 1+ \frac{Q T_0 d}{\mu}  \right)  }\,.
\end{align}

 \section{Solving Method for  \eqref{eq:opt_step2}}
 \label{appendix:solving_method}
The methodology proposed in the previous Section entails two main optimization/learning steps that need to be solved. While the solution to the optimization in Step 1 has a closed form solution, i.e., equation \eqref{eq:LSE_solution}, in Step 2 the optimization problem in \eqref{eq:opt_step2} needs to be solved efficiently. This optimization becomes computationally expensive in large-scale graphs, see Appendix C, therefore in the following we propose a computationally effective optimization algorithm. We first rewrite the problem as follows  
\begin{align}
\max_{\mat{h}} \  &\mat{D} \mat{h} + c_t\sqrt{\mat{X}\mat{V}_t^{-1}\mat{X}^T} \nonumber \\
 \text{s.t. }     	 &h(n)\in[0,1], \ \forall n \nonumber \\
			 & ||\mat{h}||_0 \leq T_0
\end{align}
where the constraints for $\mat{h}$ are explicitely related to $\mathcal{A}$, and where $\mat{D} \mat{h}=    \mat{X \alpha}$, according the notation introduced in Sec.~\ref{sec:prob_form_new}. 
Decomposing  $\mat{V}_t^{-1}$ as $\mat{V}_t^{-1} = \mat{L}^T \mat{L}$, we have  $ {\mat{X}\mat{V}_t^{-1}\mat{X}^T} = ||\mat{L} \mat{X}^T||_2^2$.  
 This leads to the following optimization problem
\begin{align}
\max_{\mat{h}} \  &\mat{D} \mat{h} + c_t ||\mat{L} * \mat{b}^T\otimes \mat{h}^T||_2 \nonumber \\
 \text{s.t. }     	 &h(n)\in[0,1], \ \forall n \nonumber \\
			 & ||\mat{h}||_0 \leq T_0
\end{align}
where we have used $\mat{X} = \mat{M}\mat{P}\mat{I}_{K+1} \otimes \mat{h}=    \mat{b}\otimes \mat{h}$.

The above problem maximizes a convex objective function over a polytope, defined by both constraints. It can be shown that if the objective function has a maximum value on the feasible region, then it is at the edges of the polytope. Therefore, the problem reduces to a finite computation of the objective function over the finite number of extreme points.
In the case of large networks, this computation could be too expensive, hence the proposed \algolight.  
\newline
\
\newline
The above problem  can   be solved also with the following approximated solution.  We relax the sparsity constraint  with a $l1$-norm constraint, leading to the following equivalent problem
\begin{align}\label{eq_appendix:l1approx}
\max_{\mat{h}} \  &\mat{D} \mat{h} + c_t ||\mat{L} * \mat{b}^T\otimes \mat{h}^T||_2 \nonumber \\
 \text{s.t. }     	 &h(n)\in[0,1], \ \forall n \nonumber \\
			 & ||\mat{h}||_1 = \sum_{n=1}^N h(n) \leq T_0
\end{align}
The   problem in \eqref{eq:l1approx} is a maximization of a convex  problem over a polytope and 
can be globally solved by   solvers for constrained nonlinear optimization problems, as  branch-and-bound search (\emph{e.g.}, bmibnb in Matlab) or    sequential quadratic programming (SQP).   The branch-and-bound solver is based on a   spatial branch-and-bound strategy, based on linear programming relaxations and convex envelope approximations. Relaxed problems are solved to evalute lowe and upper bounds. Given these lower and upper bounds, a standard branch-and-bound logic is used to select a branch variable, create two new nodes, branch, prune and navigate among the remaining nodes. On the other side, the sequential quadratic programming  is an iterative procedure which models the originating problem  by a quadratic programming  subproblem at each iteration. In both cases, the problem stays NP-hard and  the solvers require exponential computational complexity to achieve the optimal solution.  Specifically, it can be observed that the cardinality of the decision variables $|\mathcal{A}|$ is ${N \choose T_0}$ and the complexity of the solver grows exponentially with  $|\mathcal{A}|$.

\section{Details on \algolight}
\label{sec:Appendix_algo_light}
In the following, we better motivate the \algolight proposed to solve \eqref{eq:l0}.   The problem formulation in \eqref{eq:l0}  has a convex objective that needs to be maximized over a polytope. Because of the binary constraint on the $\mat{h}$ variable,  the optimal solution  will lie on one of the vertices of the boundary. Therefore, visiting all vertices would lead to the optimal solution. This however is not always practical in large graphs. Therefore, we propose an algorithm to jump from one vertex to the other in such a way that the objective function always increases and ideally the optimum point is reached within the maximum number of iteration allowed by the algorithm. 

In the following, we first proof the Lemmas needed to derive the \algolight, then we derive the partial derivative needed in Lemma 4. 

\subsection{Proof of the Lemma 3 and Lemma 4}

{\bf Lemma 3:}  \emph{One vertex is optimal if there is no better neighboring vertex.}
\newline   
\emph{Proof:} This follows directly from the definition of convex function. Namely, the point of global optimum will be the one that maximizes the objective function. Therefore, moving out in any direction from the optimum point (\emph{i.e.}, visiting any neighboring node) will not increase the objective function. At the same time, any other point which is not optimum will always have at least one neighboring node that leads to a no-worse objective function.  $\square$

\vspace{.2cm}

The missing step is how to select the best direction to move from one iteration to the other.  This is explained by the following Lemma. 

\vspace{.3cm}

\begin{figure*}
\centering
\subfigure[graphical representation]{\includegraphics[width=.35\linewidth,  draft=false]{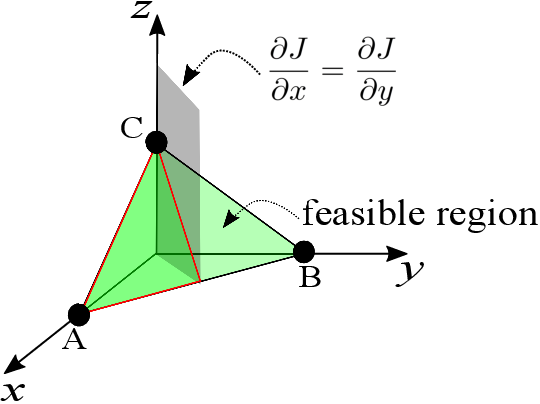}\label{fig:polotope}} \hfil
\vspace{0.1cm}
\subfigure[Geometric visualization] { \includegraphics[width=.35\linewidth,  draft=false]{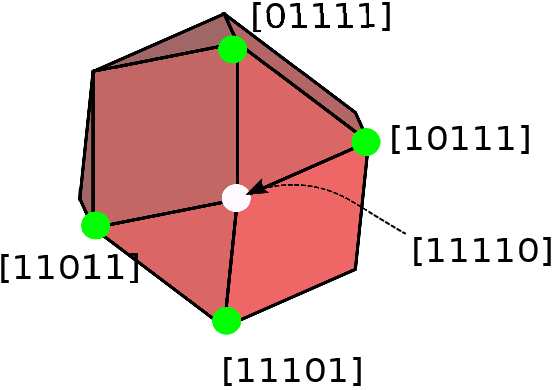}
\label{fig:polythope2}
} 
\caption{Visualization of the feasible region for a three dimensional domain. (a) Red contour delimit the area in which the gradient will be pointing to. (b) white circle denotes the starting node (or vertex). Green vertices are the candidate ones.} 
\end{figure*}

{\bf Lemma 4:}  \emph{From a vertex,   moving to one of the neighboring nodes in the direction of the greatest gradient leads to a no-worse objective function. Let denote by $h_{in}^{(i)}$  and $h_{out}^{(i)}$  be the variable that enters and leaves the set of basic variables, respectively, at the $i$-th iteration. These variables are evaluated as follows }
   $$ {in} = \arg\max_{n | h_n \notin \mathcal{B}_{t}} \left\{    \frac{\partial J}{\partial h_n} \Bigr|_{\mat{h}^{(i-1)}}   \right\}, \ \ \ \ 
    {out} = \arg\min_{n | h_n \in \mathcal{B}_{t}} \left\{    \frac{\partial J}{\partial h_n} \Bigr|_{\mat{h}^{(i-1)}}   \right\} $$
\newline   
\emph{Proof:} We first consider a simple case of $N=3$ with the three possible solutions $\mat{h}_A=[1 \, 0 \,  0]$, $\mat{h}_B=[0 \, 1 \,  0]$, $\mat{h}_C=[0 \, 0 \,  1]$, see Fig. \ref{fig:polotope}. Let also assume that $J(\mat{h}_C)<J(\mat{h}_B) <J(\mat{h}_A)$ and that $\mat{h}_C$ is the starting node. The algorithm needs to decide if moving in the direction of $\mat{h}_B$ and $\mat{h}_A$. From the definition of gradient, the gradient of $J$ evaluated in $ \mat{h}_C$ will be pointing toward $\mat{h}_A$ rather than $\mat{h}_B$ (since the improvement of the objective function if larger is moving in this direction). Let make the approximation of having a continuous domain (rather than a discrete one) highlgihted by the green shaded area in Fig.  \ref{fig:polotope} and assume to run the gradient ascend to find the optimal solution.   
The gradient from $\mat{h}_C$  will be pointing to any region highlighted in red in Fig.  \ref{fig:polotope} since at last the gradient ascend will be point toward $\mat{h}_A$. This that that the  variable at the iteration $i+1$ would be
$$\mat{h}^{(i+1)} =   \mat{h}_C +  \alpha \left(\frac{\partial J}{\partial x} \mat{i} + \frac{\partial J}{\partial x} \mat{j} + \frac{\partial J}{\partial x} \mat{k}\right).$$ 
and to move toward  $\mat{h}_A$, the following condition needs to be respected  
$$\frac{\partial J}{\partial x} \geq \frac{\partial J}{\partial y}$$

This condition can be extended to a more general case with a starting node $\mat{h}^{(i)} = [h^{(i)}_1, h^{(i)}_2, \ldots, h^{(i)}_N]$ at the $i$-th iteration. Without loss of generality we assume that the first $T_0$ variables are the basic variables, this means 
$$
\mat{h}^{(i)} = [\underbrace{h^{(i)}_1, h^{(i)}_2, \ldots,   h^{(i)}_{T_0}}_{\text{basic variables}} ,  \underbrace{h^{(i)}_{T_0+1},   \ldots,   h^{(i)}_N}_{\text{non-basic variables}} ] =  [\underbrace{ 1, 1, \ldots,   1}_{\text{basic variables}} ,  \underbrace{0,   \ldots,  0}_{\text{non-basic variables}}]
$$
 Since neighboring nodes differs in the non-basic variables only for one element, it means that only one of the  $N-T_0$ remaining  variables can become a basic variable. Let consider that the candidates neighboring vertices are  
\begin{align}
\mat{h}_a  &= [\underbrace{ 0, 1, \ldots,   1}_{T_0} ,  \underbrace{1,  0,    \ldots,  0}_{N-T_0}] \nonumber \\
\mat{h}_b  &= [\underbrace{ 0, 1, \ldots,   1}_{T_0} ,  \underbrace{0,  1,    \ldots,  0}_{N-T_0}]\nonumber \\
 \dots \nonumber \\
 \mat{h}_l  &= [\underbrace{ 0, 1, \ldots,   1}_{T_0} ,  \underbrace{0, 0     \ldots,  1}_{N-T_0}]\nonumber 
\end{align}
 This means that we simply need to select the direction out of $N-T_0$ that maximizes the gain in terms of objective function. This translates in the following condition 
\begin{align} 
 \max_{n | h_n \notin \mathcal{B}_{t}} \left\{    \frac{\partial J}{\partial h_n} \Bigr|_{\mat{h}^{(i)}}   \right\}.
 \end{align}
The above condition determines a $T_0$-dimensional plane with $T_0$possible vertices (green vertices in Fig. \ref{fig:polythope2}). To select the best one, we identify the least promising direction among the ones of the basic variables in $\mat{h}^{(i)}$. This translates in the second condition imposed in Algorithm \ref{algo_Iter}, which is 
\begin{align}
\min_{n | h_n \in \mathcal{B}_{t}} \left\{    \frac{\partial J}{\partial h_n} \Bigr|_{\mat{h}^{(i-1)}}   \right\} 
\end{align}
The motivation is the symmetric one of above, but this time we look for the least convenient direction since we look for the variable to abandon the set of basic variables. Therefore, we seek the direction that minimizes the gradient.  $\square$ 
 
\subsection{Derivation of the partial derivative}
\label{sec:Appendix_algo_light_derivative}
 The remaining step is the evaluation of the partial derivatives. We recall that the objective function is given by
$$
 J(\mat{h})= \underbrace{\mat{M}\mat{P} \mat{I}_{K+1} \otimes \mat{h} \hat{\mat{\alpha}}}_{F(\mat{h})} + c\underbrace{ ||\mat{L} * \mat{b}^T\otimes \mat{h}^T||_2}_{G(\mat{h})} \,.
$$
We now evaluate the partial derivative for each of the two components of the objective function. Namely: 
\begin{align}
\frac{\partial  F(\mat{h})}{\partial h_n} &= \frac{\partial    }{\partial h_n} \left(\mat{M}\mat{P} \mat{I}_{K+1} \otimes \mat{h} \hat{\mat{\alpha}}\right)  = \mat{M}\mat{P} \frac{\partial  }{\partial h_n} \left( \mat{I}_{K+1} \otimes \mat{h} \right)\hat{\mat{\alpha}} = \mat{M}\mat{P}    \mat{I}_{K+1} \otimes \mat{1}_n \hat{\mat{\alpha}} 
\end{align}
\begin{align}
\frac{\partial  G(\mat{h})}{\partial h_n} &= ||\mat{L} * \mat{b}^T\otimes \mat{h}^T||_2  =  =  \frac{  \left(\mat{L} * \mat{b}^T\otimes \mat{h}^T \right) }{ ||\mat{L} * \mat{b}^T\otimes \mat{h}^T||_2}   \left(\mat{L} * \mat{b}^T\otimes \mat{1}_n^T \right)^T
\end{align}
This leads to the following partial derivative
\begin{align}
\frac{\partial  J(\mat{h})}{\partial h_n} &=  \mat{M}\mat{P}    \mat{I}_{K+1} \otimes \mat{1}_n \hat{\mat{\alpha}} +   \frac{  \left(\mat{L} * \mat{b}^T\otimes \mat{h}^T \right) }{ ||\mat{L} * \mat{b}^T\otimes \mat{h}^T||_2}   \left(\mat{L} * \mat{b}^T\otimes \mat{1}_n^T \right)^T
\end{align}

\section{Additional Experimental Results}

\subsection{Performance comparison}
\label{subsection:comparison}
\begin{figure*}[t]
\centering 
\subfigure[$\tau=10$]{\includegraphics[width=.4\linewidth,  draft=false]{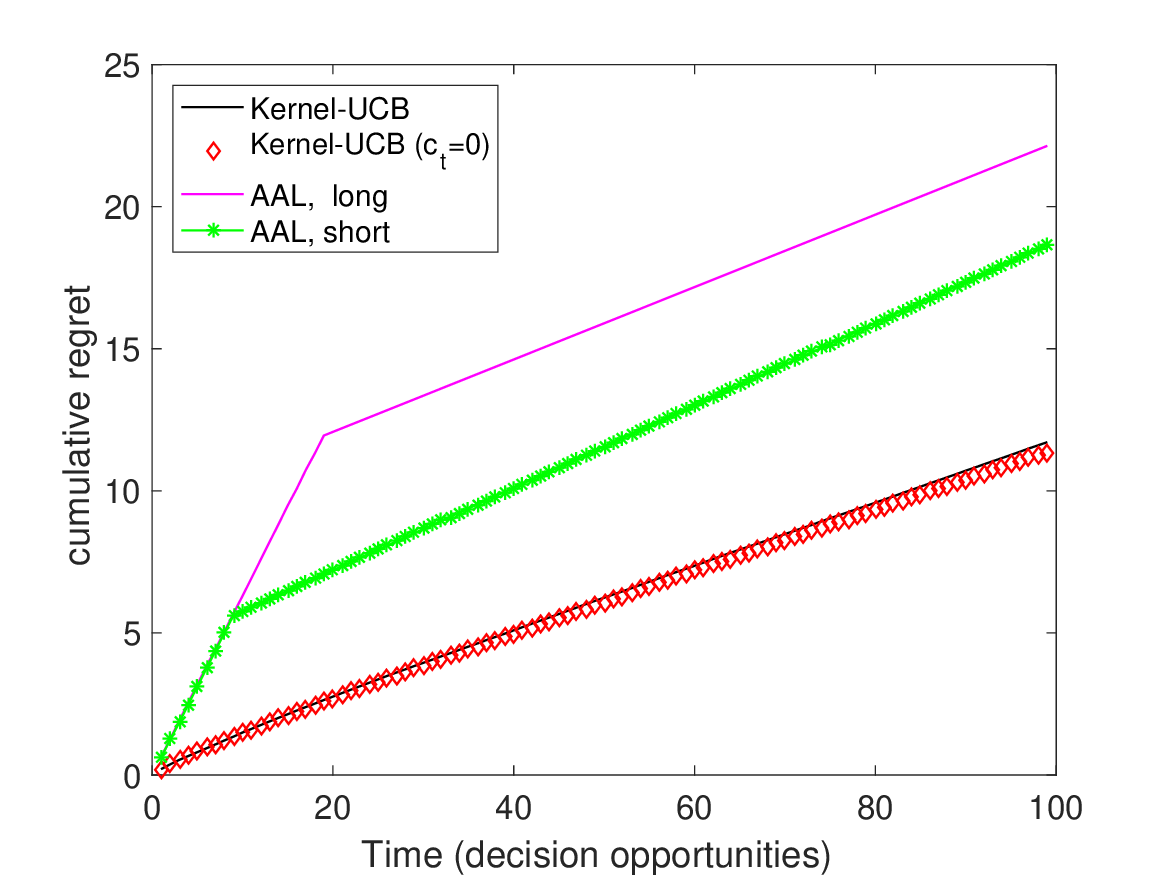}\label{fig:LoadG_Diff_Tau10_To5_Noise2}} \hfil
\vspace{0.1cm}
\subfigure[$\tau=0.5$] { \includegraphics[width=.4\linewidth,  draft=false]{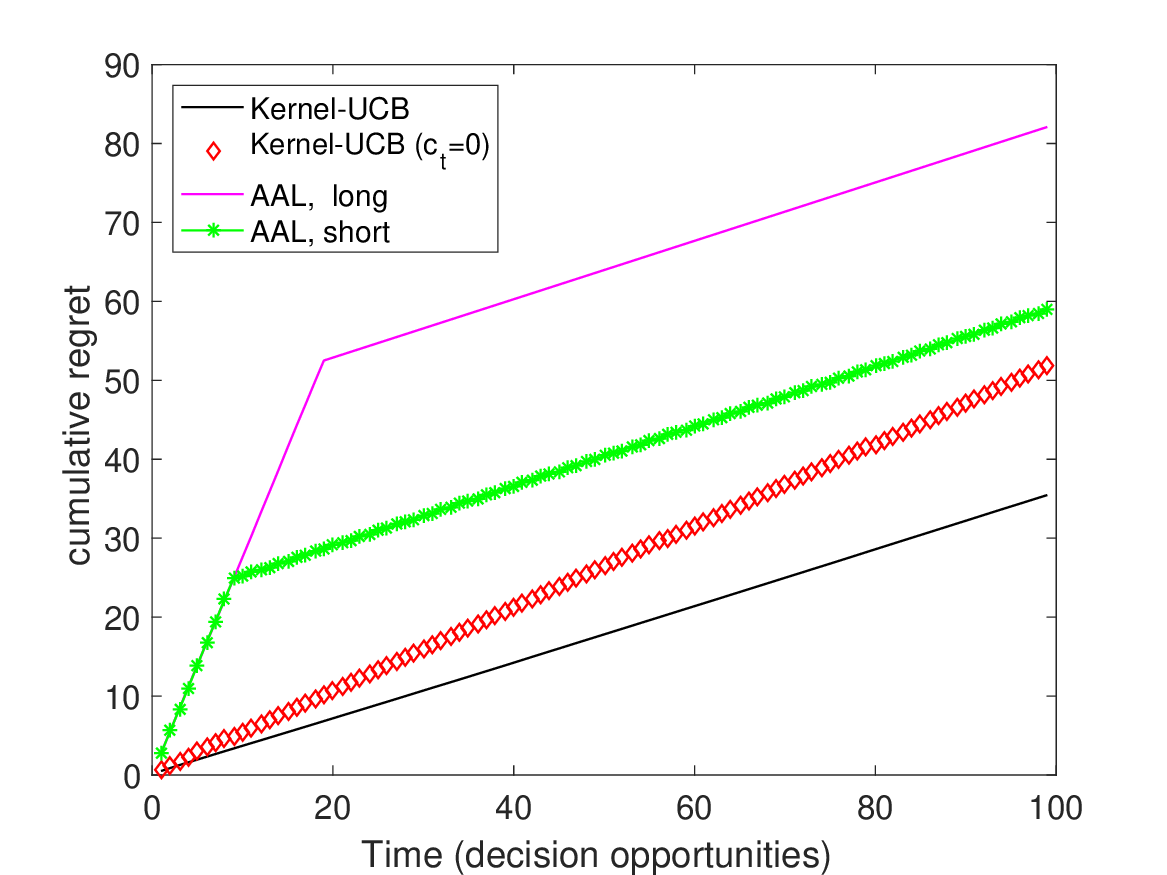}
\label{fig:LoadG_Diff_Tau05_To5_Noise2}
} 
\caption{Cumulative regret vs. time for randomly generated graphs with $N=100$, diffusion process  and sparsity level $T_0=5$.} \label{fig:LoadG_Tau05_To5_Noise2}
\end{figure*}

In Fig. \ref{fig:LoadG_Tau05_To5_Noise2}, the cumulative regret is provided as a function of time for the same settings of Fig. \ref{fig:LoadedG_Diff_T02_Noise2}, but with $\tau=10$ and  $\tau=0.5$  for the diffusion process. Larger $\tau$ means a more diffused (and therefore more informative) resulting signal. Therefore, in the case of 
$\tau=10$ the estimate of the polynomial $\mat{\alpha}$ is less challenging than the case of diffusion process with $\tau=0.5$. This can be observed by two behaviors: $1)$ a learning time of $20$ decision opportunities does not improve the estimation with respect to a learning framework of $10$ decision opportunities. This is motivated by the fact that the two AAL algorithms have the same slope of the cumulative regret. $2)$ The cumulative regret evaluated with the Kernel-UCB with $c_t=0$ almost overlaps with the curve obtained from  Kernel-UCB with confidence bound. This means that taking into account the uncertainty of the estimation is less beneficial in Fig. \ref{fig:LoadG_Diff_Tau10_To5_Noise2}, and this is due to the better estimate of the polynomial. On the contrary, in the case of  a more localized process, namely $\tau=0.5$, the estimation process is more challenging, hence taking into account the confidence bound leads to a better selection of the actions $\mat{h}$ over time.

\subsection{Influence of the Graph Topology }

We are now interested in understanding how much the graph topology correlates to the performance of the learning process. To do this,  we first consider a randomly generated training set of $300$ signals, and we then estimate the accuracy of the learned polynomial $\alpha$. Implicitly, the better the estimate, the better the decision maker behaviour. To measure the accuracy of the estimate, we evaluate the error on the resulting signal given the action $\mat{h}$ of test signals. Basically we evaluate $(1/ |Y_{Test}|)\sum_i || \mat{y}_i - \mathcal{D} \mat{h}_i||_2^2 / ||\mat{y}_i||^2$, where $|Y_{Test}|$ is the cardinality of the testing set. 
  \begin{figure*}[t]
\centering 
\subfigure[High connected graph (T=0.95)]{\includegraphics[width=.4\linewidth,  draft=false]{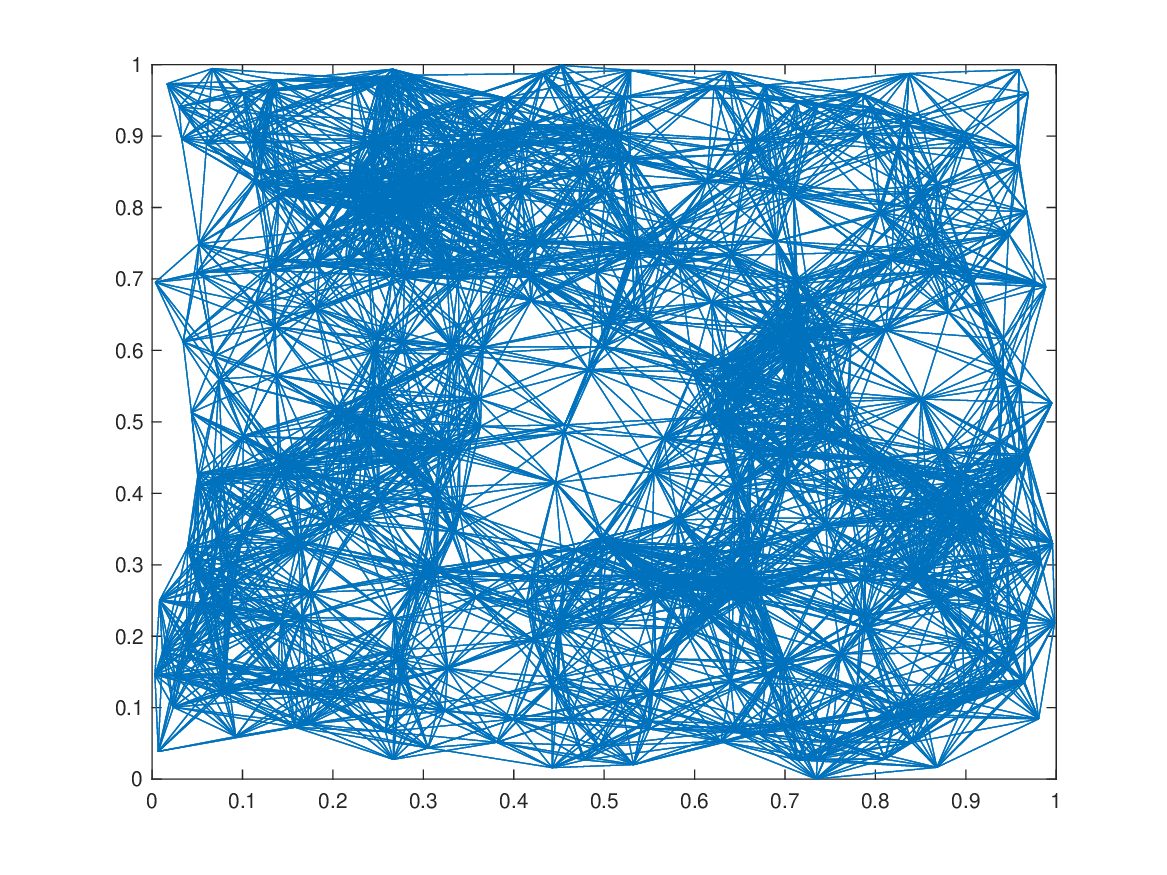}\label{fig:graph_T095}} \hfil
\vspace{0.1cm}
\subfigure[Low connected graph (T=0.987)] { \includegraphics[width=.4\linewidth,  draft=false]{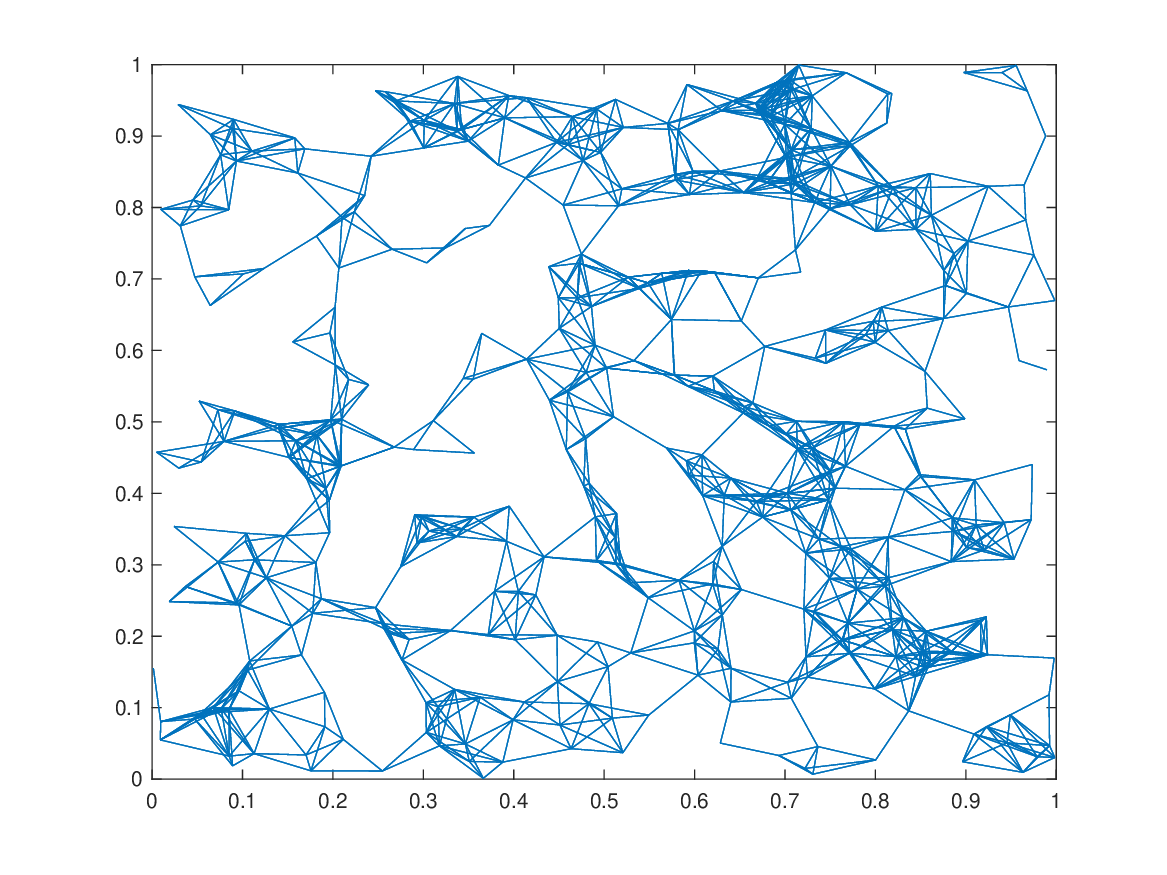}
\label{fig:Graph_0987}
} 
\caption{Different graph topologies in the case of $N=400$ nodes, $\sigma = 0.5$ and we vary $T=0.95$ and $T=0.987$.} \label{fig:graphN400}
\end{figure*}

\begin{figure*}[t]
\centering
\subfigure[Graph Laplacian Eigenvalues]{\includegraphics[width=.4\linewidth,  draft=false]{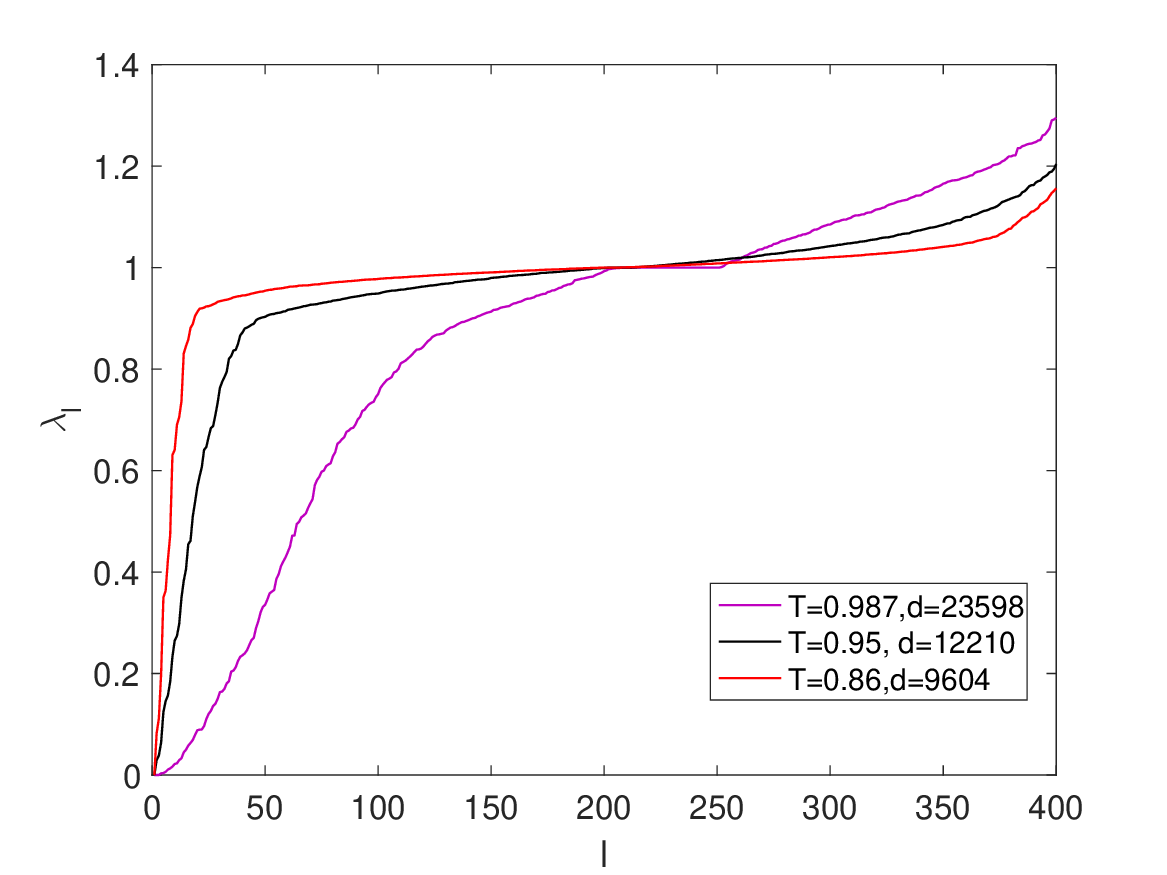}\label{fig:Laplacian_Comparison_N400}} \hfil
\vspace{0.1cm}
\subfigure[Estimation Error] { \includegraphics[width=.4\linewidth,  draft=false]{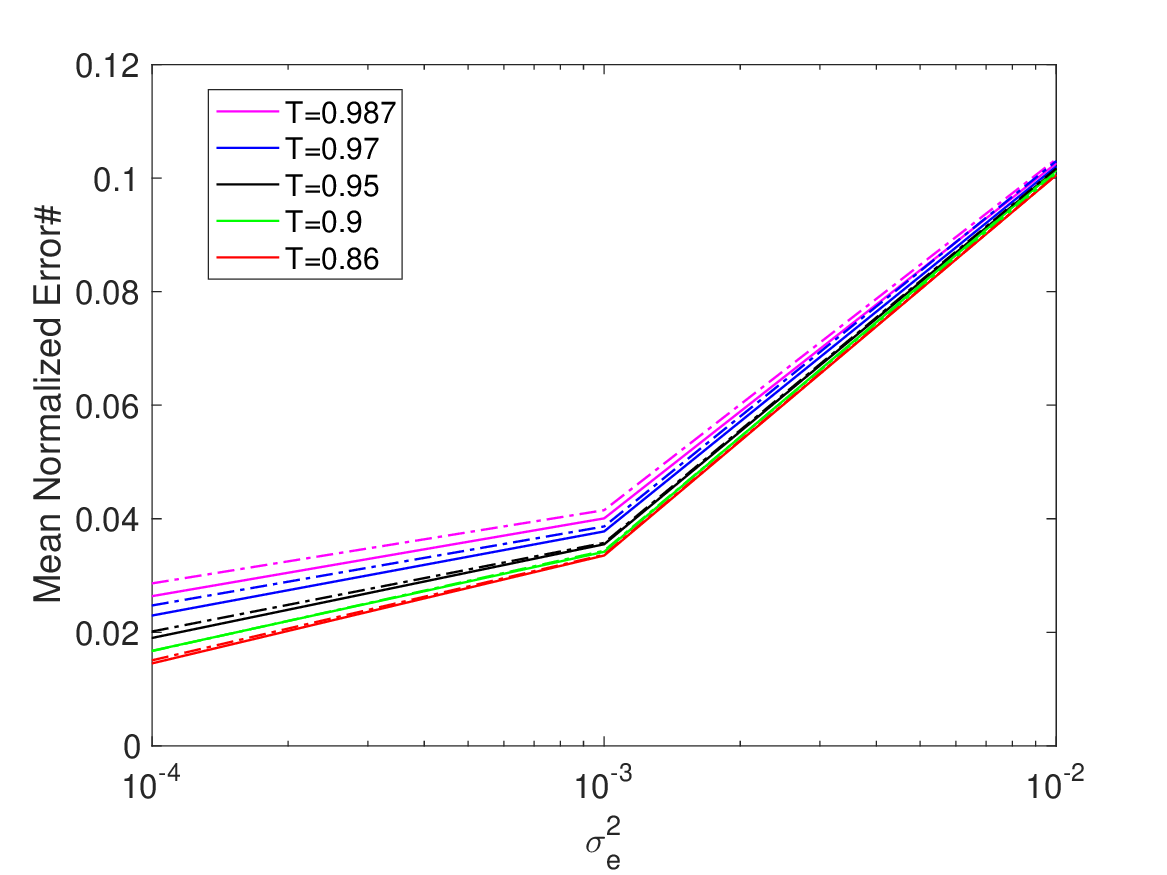} \label{fig:Error_vs-Topologies}} 
\caption{Graph Laplacian distribution and signal estimation error for random graphs with different levels of connectivity,  $N=400$ nodes, and sparsity value $T_0=25$.The estimation error is evaluated both in the case of full observability (solid line) or in the case of partial observability (dotted line) with a mask covering $40\%$ of the nodes. } 
\end{figure*}

\begin{figure}[t]
\centering
\includegraphics[width=.6\linewidth,  draft=false]{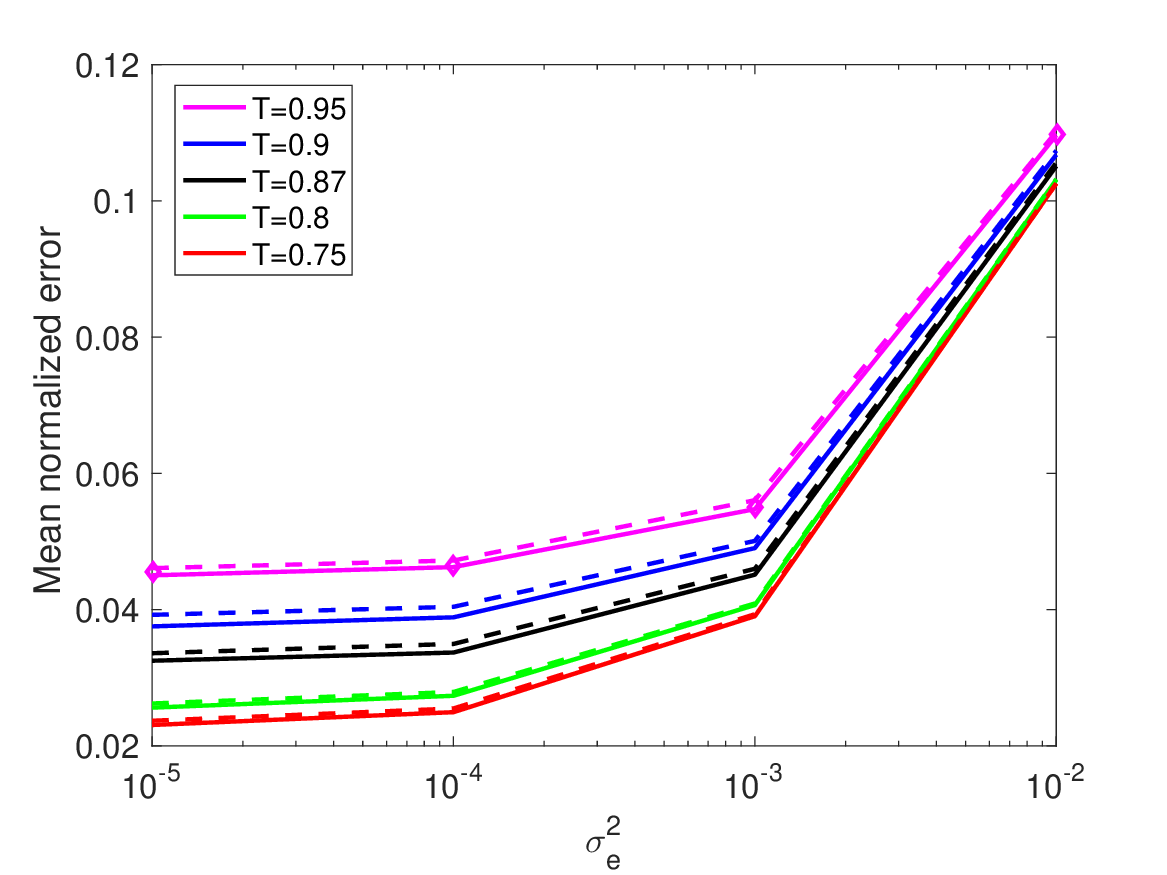}
\caption{Signal estimation error vs noise variance for random graphs with different levels of connectivity,  $N=100$ nodes, and sparsity value $T_0=15$.The estimation error is evaluated both in the case of full observability (solid line) or in the case of partial observability (dotted line) with a mask covering $40\%$ of the nodes. } \label{fig:N100_T015_errorVSnoise} 
\end{figure}

The key intuition we have from Lemma 2 is that the confidence bound (and therefore the uncertainty on the estimation) increases with  the sparsity level $T_0$ as well as the parameter $d$, which represents the power sum of the eigenvalue of the Laplacian\footnote{Note that the estimation error increases also with variance of the random noise $\mat{\eta}$ but this is not related to the graph topology, therefore it is beyond the scope of this section.}.  Specifically, 
\begin{align}
d &=\sum_{k=0}^K \sum_{l=1}^N \lambda_l^k \leq \tilde{N}  \sum_{k=0}^K  \lambda_{\text{max}}^k = \tilde{N} \frac{1-\lambda_{\text{max}}^{k+1}}{1-\lambda_{\text{max}}}
\end{align}
where $\tilde{N}$ is the number of eigenvalues considerably larger than $0$, and $\lambda_{\text{max}}$ the maximum eigenvalue of the graph Laplacian. Note that the last equality holds from the geometric series. 

We are therefore interested in studying how much the estimation error depends on the connectivity of the graph, and therefore on the Laplacian. We consider graphs generated with the RBF model. By changing the threshold parameter $T$, we generate more or less connected graphs. In Fig. \ref{fig:graphN400}, the graph topologies for $T=0.95$ and $T=0.987$ when $N=400$ is provided. As expected, the smaller   $T$ is,  the more connected is the graph. Higher levels of connectivity also leads to a more narrow profile of the eigenvalues of the Laplacian $\lambda_l$, as observed from Fig. \ref{fig:Laplacian_Comparison_N400}, where the values of  $\lambda_l$ are provided for different graph topologies.  In particular, we provide $\lambda_l$ for graph topologies with $N=400$ and $T=0.987, 0.95$ and $0.86$. In the legend, we also provide the power sum of the eigenvalues, namely $d$, for each graph topology. As a consequence, more connected graphs lead to a more accurate estimate of the generating kernels, see Fig. \ref{fig:Error_vs-Topologies}. The intuition is that more connected graphs lead to a more informative resulting signal $\mat{y}$  and therefore to a better estimate. Mathematically, this can also be deduced by observing the distribution of the Laplacian eigenvalues (Fig. \ref{fig:Laplacian_Comparison_N400}) and the associated power sum $d$. 
Similar behavior is observed in the case $100$ nodes, as depicted in Fig. \ref{fig:N100_T015_errorVSnoise}.


\begin{figure}[t]
\centering
\includegraphics[width=.6\linewidth,  draft=false]{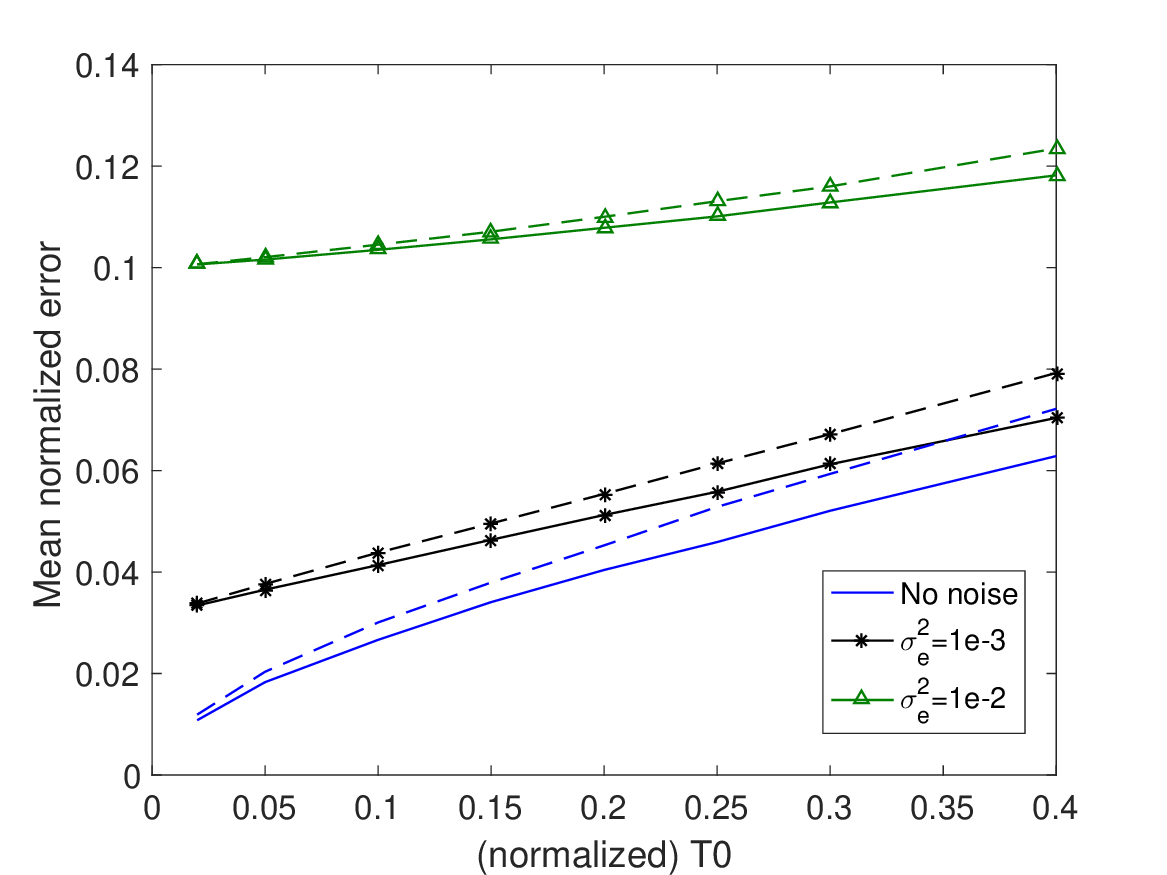}
\caption{Mean normalized error vs the sparsity level of the source signal for graphs with $100$ (dashed line) and $400$ (solid line) nodes  in the case of full observability.} \label{fig:N400_100_Vs_T0} 
\end{figure}

\subsection{Influence of source sparsity}
We are now interested in studying the impact of sparcity level on the uncertainty bound, since the theoretical bound shows an explicit dependency. 
In Fig.  \ref{fig:N400_100_Vs_T0}, we then evaluate the mean normalized error as a function of the sparsity level of the source signal.   Results are plotted as a function of the normalized sparsity level, defined as $T_0/N$, and they are provided  for a random graph with  $100$ nodes (dashed line) and with $400$ nodes (solid line). Finally, a diffusion process with $\tau=10$ is considered and estimated by the polynomial $\mat{\alpha}$ of degree $K=20$. Simulations are considered for different noise variances, namely $10^{-3}$ and $10^{-2}$, which are compared to the noiseless case.  It is interesting to observe that larger sparsity levels lead to an increase of the estimation error. This is motivated by the following observation: less source signals (\emph{i.e.}, smaller sparsity value $T_0$) lead to more localized information so that sources are more informative. Therefore, a more sparse signal $\mat{h}$ leads to a better estimation.   This can also be demonstrated by Lemma 2, where mathematically it can be observed that the confidence bound is proportional to the sparsity value $T_0$. Therefore the higher $T_0$ the greater the uncertainty about the estimation. Finally, from Lemma 2 the estimation error should increases also with the number of nodes $N$, while in Fig.  \ref{fig:N400_100_Vs_T0} we observe the opposite behavior. This is because the number of  nodes increases but within the same unitary space (varying the density of the graph) and it is observed that denser graphs ($N=400$) outperform less dense ones ($N=100$). This is in line with the above observation that more connected (and more dense) graphs lead to a better estimate. 

\end{document}